%% file: hu.tex
\documentstyle[a4wide,url,afd-gen,afd-psm,afd-prog,afd-eis,pict-gen,pict-er,pict-hyp,
               pict-act,pict-psm,times,gu,biblio]{article}

\begin{document}
   \input{hutail}

   \BIBLIOGRAPHY{alpha}
\end{document}

%% file: hutail.tex
\title{
   A  General Theory for
   the Evolution of Application Models
}

\author{
   H.A. Proper$^1$ and
   Th.P. van der Weide\\[0.3cm]
   Computing Science Institute,
   University of Nijmegen\\
   Toernooiveld,
   NL-6525 ED Nijmegen,
   The Netherlands\\
   E.Proper@acm.org
}
\footnotetext[1]{Currently at:
                Department of Computer Science, University of Queensland,
                Queensland 4072, Australia}

\date{}

\maketitle

\NoFloatPages
\input{gu00}  

\input{pub}

\begin{quote}
\end{quote}

\input{hu01}  
\input{hu02}  
\input{hu03}  
\input{gu04}  
\input{gu05}  
\input{hu06}  
\input{hu07}  
\input{hu08}  
\FloatPages

%% file: gu00.tex
\begin{abstract}
   In this article we focus on evolving information systems.
   First a delimitation of the concept of evolution is provided, 
   resulting in a first attempt to a general theory
   for such evolutions.

   The theory makes a distinction between the underlying
   information structure at the conceptual level, its evolution 
   on the one hand,
   and the description and semantics of operations on the information
   structure and its population on the other hand.
   Main issues within this theory are object typing,
   type relatedness and identification of objects.
   In terms of these concepts, we propose some axioms on the well-formedness
   of evolution.

   In this general theory,
   the underlying data model is a parameter,
   making the theory applicable for a wide range of modelling techniques,
   including object-role modelling and object oriented techniques.
\end{abstract}

%% file: pub.tex
{\sc Published as:}
\begin{quote}
  H.A.~(Erik) {Proper} and Th.P. van~der {Weide}. {A General Theory for the Evolution of Application Models -- Full version}. Technical report, Department of Computer Science, University of Queensland, Brisbane, Australia, 1994.
\end{quote}

%% file: hu01.tex
\section{Introduction}

As has been argued in \cite{Article:91:Roddick:Evolution}
and \cite{Report:92:Falkenberg:EISPlea},
there is a growing demand for information systems,
not only allowing for changes of their information base,
but also for modifications in their underlying structure
(conceptual schema and specification of dynamic aspects).
In case of snapshot databases, structure modifications
will lead to costly data conversions and reprogramming.

The intention of an evolving information system 
(\cite{Report:91:Falkenberg:EISConcepts},
 \cite{Report:92:Falkenberg:EISPromo}) 
is to be able to handle updates of all components of the so-called 
{\em application model},
containing the information structure,
the constraints on this structure,
the population conforming to this structure
and the possible operations.
The theory of such systems should, however,
be independent of whatever modelling technique is used to describe
the application model.
In this paper, we discuss a general theory for the
evolution of application models.
However, only conceptual aspects are considered, focus is on {\em what}
evolution is, rather than on {\em how} to implement evolution in a
database manegement system.
In \cite{Report:92:Proper:TowardsEvolvAM}, an informal introduction
to this theory is provided.

The central part of this theory will make weak assumptions on the 
underlying modelling technique,
making it therefore applicable for a wide range of
data modelling techniques such as
ER (\cite{Article:76:Chen:ER},
    \cite{Book:89:Yourdon:Analysis}),
EER (\cite{Report:90:Engels:ConcMod}),
NIAM (\cite{Book:90:Nijssen:NIAM}),
IFO (\cite{Article:87:Abiteboul:IFO})
and the generalized object role data modelling technique
PSM (\cite{Report:91:Hofstede:PSM},
     \cite{Report:91:Hofstede:PSMPromo}),
action modelling techniques such as
Task Structures (\cite{Report:91:Wijers:KnowlMod},
                 \cite{Report:91:Hofstede:TaskAlg}),
DFD (\cite{Report:89:Bruza:DataFlowSem})
and ExSpect (\cite{Article:89:Hee:Exspect}),
and furthermore object oriented modelling techniques
(\cite{Article:90:Korson:OOParadigma})
adhering to the OO typing mechanism as described in
\cite{Article:85:Cardelli:DBTypeTheory}.
In \cite{Report:93:Proper:EvolvPSM}, the application of the theory
presented in this article to the object-role modelling technique
PSM, leading to EVORM, is described.
Some of the reasons for choosing an object-role modelling technique
as a first application of the general theory, are the steadily
increasing popularity of object role modelling
(\cite{Book:93:Nijssen:NIAM-ISDM}, \cite{Book:94:Halpin:ORM-1}),
and the existence of a formal definition of object-role modelling.
An further advantage of object-role modelling is that it is supported
by an effective, natural language based, conceptual schema design 
procedure (way of working).
The extension of this technique to cover evolution aspects are
not covered by this paper.

The assumptions suppose a typing mechanism for objects,
a type relatedness relation expressing which object types may share
instances,
and a hierarchy on object types expressing inheritance of
identification.

In \cite{Article:90:Snodgrass:TempDBStatus} a classification for
incorporating time in information
systems (databases) is presented.
This classification makes a distinction between 
rollback, historical and temporal information systems (databases).
However, all these classes do not yet take schema evolution into account.
For this reason, we propose a new class: evolving information systems.

We mention some examples of research regarding these first three
classes.
In the TEMPORA project
(\cite{Article:91:Theodoulidis:Tempora},
\cite{Article:92:McBrien:Tempora}),
the ER model is enhanced with the notion of time,
resulting in the ERT model.
In TODM (\cite{Article:86:Ariav:TemporalDataMod}) and
ERAE (\cite{Report:85:Dubois:ERAE}),
similar strategies are followed, extending the relational model
with the notion of time.
This makes it possible to handle historical data,
over a (non-varying) underlying information structure.
In \cite{Article:88:Saake:DynConstr},
\cite{Article:91:Saake:ObjectBehaviour}
and \cite{PhdThesis:88:Saake:Objects}
the focus is on the monitoring of dynamic constraints,
i.e., constraints over such historical data.
Dynamic constraints restrict temporal evolutions,
i.e., state sequences of databases.
Historical data, however,
are considered in their approach only as a means for implementing a
monitor.
Only the object domains may vary in the course of time.

Within the class of evolving information systems, 
extensions of object oriented modelling techniques with a
time dimension (both on instance and type level) can be seen as 
a first subclass, providing first results on the actual implementation
of evolution of schemas in database management systems.
In \cite{Article:86:Skarra:OOEvolution} a taxonomy for type evolution
in object oriented database management systems is provided.
The ORION project (\cite{Article:87:Banerjee:OOSchemaEvol},
\cite{Article:89:Kim:Orion})
offers a more detailed taxonomy, together with a (semi formal) semantics
of schema updates restricted to object oriented databases.
The ORION system, together with the GemStone
(\cite{Article:87:Penney:OOEvolution}, \cite{Article:89:Bretl:Gemstone}),
and \keyword{Sherpa} (\cite{Article:89:Nguyen:OOSchemaEvol}) systems,
are among the first object oriented database systems to support 
schema/type evolution.
In the \keyword{Cocoon} project (\cite{Article:91:Tresch:OOEvolution},
\cite{Article:92:Tresch:OOEvolution}) an approach to the evolution of 
schemas in object oriented databases is followed in which schema objects 
(e.g.\ object types) are considered to be objects like others (from the 
application).
We will do a similar thing, and consider objects of both levels as objects
describing an evolution in the course of time.
This paper differs from these object oriented database management
systems, in that the paper focusses on the underlying concepts of
evolution rather than implementation of these concepts.

The second subclass of evolving information systems can be found in the
field of version modelling, which can be seen as a restricted form of 
evolving information systems
(\cite{Article:90:Katz:VersionModel},
 \cite{Article:90:Mylopoulos:Telos},
 \cite{Article:92:Jarke:DAIDA}).
An important requirement for evolving information systems,
not covered by version modelling systems,
is that changes to the structure can be made on-line.
In version modelling,
a structural change requires the replacement of the old system
by a new system, and a costly conversion of the old population into a
new population conforming to the new schema.
Experiences in the field of version modelling can be fruitfully employed
when actually implementing an evolving information system (management
system).

A third subclass of research regarding evolving information systems
extends a manipulation language for relational models with historical
operations,
both on population and schema level.
An example of this approach can be found
in \cite{Article:90:McKenzie:Evolution},
in which an algebra is presented
allowing relational tables to evolve by changing their arity.
This direction is similar to the ORION project
(\cite{Article:87:Banerjee:OOSchemaEvol}, \cite{Article:89:Kim:Orion}),
in that a manipulation language is extended with operations supporting
schema evolution.

\Fig[\FrameTechn]{Framework for methodologies}
In this paper we consider evolving information systems,
and try to abstract from the subclasses mentioned above.
Therefore,
we take the underlying informaton structuring technique for granted,
make only weak assumptions on the underlying technique, and limit ourselves
to conceptual issues.
This paper restricts itself basically to the {\em way of modelling}
in the framework for methodologies of \SRef{\FrameTechn}.
This framework, taken from~\cite{PhdThesis:91:Wijers:ModellingSupport},
presents a more structured view on methodologies,
and is based on the original framework
of~\cite{Article:89:Seligmann:Framework}.
It makes a distinction between a way of thinking, a way of control,
a way of modelling, a way of working and a way of support.
The way of thinking is concerned with the philosophy
behind the methodology and contains basic
assumptions and viewpoints of this methodology. The way
of control captures project management. 
The way of modelling describes the models and model
components used in the methodology, while the way of working
describes strategies and procedures of how to arrive 
at specific models. 
The way of support, finally, is concerned with technique/method support.
The GemStone, ORION, Sherpa and Cocoon systems can thus be seen
as first attempts for a way of support for evolving information systems.
However, to our knowledge, all these systems lack a rigourously
formalised underlying way of modelling. 
Although it is benificial to have a running way of support as soon
as possible, having a well thought out underlying way of modelling
first has proven its usefullness.
At least, this should be the second goal after completing the tool!

The structure of the paper is as follows.
In \SRef{Approach} we describe the approach that has been taken
to the concept of evolution,
in which evolution is seen (similar as history books) as an ensemble
of individual histories of application model elements.
As we will not focus on a particular modelling technique,
\SRef{AppMod} describes the
minimal requirements for an underlying technique,
as discussed above.
In \SRef{AppModVersion} we introduce the universe for application model
evolution.
After that,
we discuss what constitutes a wellformed application model version.
In \SRef{AMEvolv} the evolution of application models is treated,
and some wellformedness rules for such evolutions are formulated.

%% file: hu02.tex
\section{An Approach to Evolving Information Systems}
\SLabel{section}{Approach}

In this section we discuss our approach to evolving information
systems.
We start with a hierarchy of models, which together constitute
a complete specification of (a version of) a universe of discourse
(application domain).
Using this hierarchy, we are able to identify that part of an 
information system that may be subject to evolution.
From this identification,
the difference between a traditional information system,
and its evolving counterpart, will become clear.
This is followed by a discussion on how the evolution of an information
system is modelled.

\input{gu0201} 
\input{hu0203} 
\input{hu0204} 
\input{hu0205} 
\input{hu0206} 
\input{eu0207} 

%% file: gu0201.tex
\subsection{The extent of the corpus evolutionis}

According to \cite{Report:82:ISO:Concepts},
a conceptual (i.e. complete and minimal) specification
of (a version of) a universe of discourse consists of the following components:
\begin{enumerate}
\item an \Keyword{information structure},
      a set of \Keyword{constraints}
      and a \Keyword{population} conforming to these requirements.
\item a set of \Keyword{action specifications}
      describing the transitions that can be performed by the system.
\end{enumerate}
\Fig[\ModelsHierarchy]{A hierarchy of models}
The set of action specifications in such a specification 
is referred to as the \Keyword{action model}.
The \keyword{action model} describes all possible transitions on 
populations, and is usually (as good as possible) modelled by means of Petri-net like
specifications (such as ExSpect or Task Structures),
or languages based on SQL.
The \Keyword{world model} encompasses the combination of information 
structure, constraints and population.
A conceptual specification of a universe of discourse,
containing both the action and world model,
is called an \Keyword{application model}
(\cite{Report:91:Falkenberg:EISConcepts}, 
 \cite{Report:92:Proper:TowardsEvolvAM}).
The resulting hierarchy of models is depicted in 
\SRef{\ModelsHierarchy}.

The part of an (evolving) information system that is allowed to change
over time, will be referred to as \Keyword{corpus evolutionis}.

\Comment{
Conform \Keyword{corpus delicti}, meaning the \Keyword{substance of crime},
corpus evolutionis is the \Keyword{substance of evolution}.
} 

In most traditional information systems, the corpus evolutionis 
is restricted to the population.
Nevertheless, some traditional information systems do support
modifications of other components from the application model,
to a limited extend.
For example, adding a new table in an SQL system is easily done.
However, changing the arity of a table, or some of its attributes, will 
result in a time consuming table conversion,
which also leads to loss of the old table!
In an evolving information system, the entire application model is allowed
to evolve on-line, while keeping track of the entire history of the
application model.
As a result, no information is lost in the course of history.
Note that the information capacity (introduced in 
\cite{Article:86:Hull:InfoCapacity}) may vary in the course of time.

The application model can then be looked upon as the formal denotation
of the corpus evolutionis, and 
is denoted in terms
of object types, constraints, instantiations, action specifications,
etc.
As a collective noun for these modelling concepts
the term \Keyword{application model element} is used.
Thus, in an evolving information system,
the complete application model,
described as a set of application model elements, is allowed to change
in the course of time.

%% file: hu0203.tex
\subsection{An example of evolution}
\SLabel{subsection}{RunningExample}

As an illustration of an evolving universe of discourse, consider a rental
store for audio records (LP's).
In this store a registration is maintained of
the songs that are recorded on the available LP's.
In order to keep track of the wear and tear of LP's, the number of times 
an LP has been lent, is registered.
The information structure and constraints of this universe of discourse
are modelled in \SRef{\LPLibrary} in the style of ER,
according to the conventions of~\cite{Book:89:Yourdon:Analysis}.
Note the special notation of attributes (\LISA{Title})
using a mark symbol (\#) followed by the attribute (\# \LISA{Title}).
\Fig[\LPLibrary]{The information structure of an LP rental store}

An action specification in this example is the rule \LISA{Init-freq},
stating that whenever a new LP is added to the assortment of the store,
it's lending frequency must be set to $0$:\\
~~\\
   \LISA{ACTION Init-freq} =\\
   \LISA{~~~ WHEN ADD Lp:$x$ DO}\\
   \LISA{~~~~~~ ADD Lp:$x$ has Lending-frequency of Frequency:$0$}\\
~~\\
This action specification is in the style of LISA-D
(\cite{Report:91:Hofstede:LISA-D}).
Note that the keyword `\LISA{has}' connects object types to relation types,
and the keyword `\LISA{of}' just the other way around.

After the introduction of the compact disc, and its conquest of a sizeable 
piece of the market,
the rental store has transformed into an `LP and CD rental store'.
This leads to the introduction of the object type Medium as a common
supertype (denominator) for LP and CD.
This makes CD and LP to subtypes of Medium.
Note that some modelling techniques (ECR, PSM, IFO) also feature a
construct allowing for the introduction of generalised (polymorphic)
types.
However, since LPs and CDs are identified by the same properties
(Title and Artist), in this case a subtyping relation is the most appropriate
solution.
The relation type Medium-type effectuates the subtyping of
Medium into LP and CD. 
In the new situation, the registration of songs on LP's is extended to cover
CD's as well.
The frequency of lending, however, is not kept for CD's, as CD's are 
hardly subject to any wear and tear.
As a consequence,
the application model has evolved to \SRef{\LPCDLibrary}.
This requires an update of the typing relation of instances of
object type LP,
which are now instances of both LP and Medium.
Note that this modification can be done automatically.
\Fig[\LPCDLibrary]{The information structure of a LP and CD rental store}

The action specification \LISA{Init-freq} evolves accordingly,
now stating that whenever a medium is added to the assortment
of the rental store,
it's lending frequency is set to $0$ provided the medium is an LP:\\
~~\\
   \LISA{ACTION Init-freq} =\\
   \LISA{~~~ WHEN ADD Medium:$x$ DO}\\
   \LISA{~~~~~~ IF Lp:$x$ THEN}\\
   \LISA{~~~~~~~~~ ADD Lp:$x$ has Lending-frequency of Frequency:$0$}\\
~~\\
After some years, the CD's have become more popular than LP's.
Consequently, the rental store has decided to stop renting LP's and to 
become a CD rental store.
Besides,
the recording quality of songs on CD's has appeared to be relevant
for clients.
As this quality may differ from song to song on a single CD,
and may for some song be different for recordings on different CD's,
the recording quality is added as a (mandatory) attribute to the Recording relation.

This change in the rental store, leads to the information structure
as depicted in \SRef{\CDLibrary}.
As a result of this evolution step, the action specification \LISA{Init-freq}
can be terminated, since the lending frequency of CD's is not recorded
anymore.
Furthermore,
the addition of the mandatory attribute Quality enforces an update
of the existing population.
In this case, contrary to the previous evolution step,
information has to be added to the old population.
This could, for example, be effectuated by the following
transaction:\\
~~\\
  \LISA{ADD TO Recording MANDATORY ATTRIBUTE Quality;}\\
  \LISA{UPDATE Recording SET Quality = 'AAD'}\\
\Fig[\CDLibrary]{The information structure of a CD rental store}

%% file: hu0204.tex
\subsection{The approach}

The three ER schemata, and the associated action specifications, as discussed 
above, 
correspond to three distinct snapshots of an evolving universe of discourse.
Several approaches can be taken to the modelling of this evolution.
A first approach is to model the history of application model elements by 
adding birth-death relations to all object types in the information structure
(\cite{Article:91:Theodoulidis:Tempora}),
as illustrated in \SRef{\ExplicitTimeExmpl}.

\Fig[\ExplicitTimeExmpl]{Adding history explicitly}
This approach, however, is very limited,
as it only enables the modelling of evolution of the population of
an information system.
For example, the evolution of the Recording relation type can not be modelled
in this approach.
Evolution of other application model elements than from the population,
must then be described by using a meta modelling approach
(\cite{Article:85:Snodgrass:TimeTaxonomy}).

\Fig[\SnapShots]{Evolution modelled by snapshots}
This paper takes another approach,
and treats evolution (or rather the time axis) of an application model
as a separate concept.
This still makes it possible to derive the view on the evolution of the
application model of the
first approach. 
It will, in particular, still be possible to derive the view on the
evolution of the population as presented in \SRef{\ExplicitTimeExmpl}.
Note that this approach has a resemblance to the approach from
\cite{Article:86:Skarra:OOEvolution},
which, however, is more restricted in the sense that is 
more directed towards an implementation.

When taking the second approach, there still are two alternatives to deal 
with the history of application models.
The first one is to maintain a version history of application models
in their entirety.
This alternative leads to a sequence of snapshots of application models,
as illustrated in \SRef{\SnapShots}.
The second alternative, is to keep a version history per element, thus 
keeping track of the evolution of individual object types, instances, 
methods, etc.
This has been illustrated in \SRef{\ElementEvolution}.
Each dotted line corresponds to the evolution of one distinct element.
\Fig[\ElementEvolution]{Evolution modelled by functions over time}

The major advantage of the second alternative is that it enables one to 
state rules about, and query, the evolution of distinct application model 
elements.
The first alternative clearly does not offer this oppertunity,
as it does not provide relations between successive versions of the
application model elements.

Furthermore, the snapshot view from the first alternative can be derived
by constituting the application model version of any point of time from the 
current versions of its components (consequently the view on the evolution
of populations of the first approach can be derived as well).
This derivation is examplified in \SRef{\SnapShotEvolution}.
In the theory of evolving application models we will therefore adapt the
second alternative.
\Fig[\SnapShotEvolution]{Deriving snapshots from element evolutions}

The above discussion has a parallel in the description of the history
(evolution) of the world.
Many approaches are possible.
One may choose to describe the evolution of the world as a sequence of snapshots,
where each snapshot contains all facts valid at a given point in time.
This (higly inefficient) way of describing the history of the world is not
used in practice, as one always wants to know the distinct evolution of 
persons, municipalities, families, laws, countries, etc.
Consequently, if one wants to make a complete description of the evolution
of the world, the evolutions of all relevant components of the world
(persons, mountains, tectonic plates, \ldots) have to be described
separately. 
Finally, we realise that the approach we take to the evolution of
application models is not new.
The described approach is in line with approaches discussed in e.g.
\cite{Article:86:Skarra:OOEvolution},
\cite{Article:87:Banerjee:OOSchemaEvol},
and \cite{Article:90:Katz:VersionModel}.
However, in this article we try to use this approach as a corner stone
for a theory of application model evolution that abstracts as much as
possible from underlying concrete modelling techniques and from
implementation related details. 
It is this theory that is the main contribution of this article. 
The aim of the theory is not to reject or replace any of the existing  
approaches to schema evolution,
but rather to complement it and provide a more elaborate theoretical 
background.

%% file: hu0205.tex
\subsection{Evolving information systems}
\SLabel{subsection}{EIS}

We are now in a position to formally introduce evolving information systems.
The intention of an evolving information system is to describe
an \Keyword{application model history}.%
\footnote{In this paper, the difference between recording and event 
time \cite{Article:85:Snodgrass:TimeTaxonomy}, and the ability to
correct stored information are not taken into consideration.
For more details, see
\cite{Report:91:Falkenberg:EISConcepts} or
\cite{Report:92:Falkenberg:EISPlea}.}
An application model history in its turn, is a set of 
\Keyword{(application model) element evolutions}[application model element evolution].
Each element evolution describes the evolution of a specific application model element.
An element evolution is a partial function assigning to points of time the actual 
occurrence (version) of that element.

An example of an \keyword{element evolution}[application model element
evolution] is
the evolution of the relation type named \LISA{Recording} in the
rental store.
When CD's are added to its assortment, the
version of the application model element \LISA{Recording} changes from
a relation type registrating songs on LP's,
to a relation type registrating songs on Media.
The removal of LP's from the assortment leads to the change of the
application model element \LISA{Recording} into a relation type registrating
songs on CD's.

The domain $\AMH$ for application model histories
is determined by the following components:
\begin{enumerate}
   \Intro{\AMH}{application model histories}
   \item The set $\AME$ is the domain for the evolvable elements of an
         application model.
         A formal definition of $\AME$ will be provided in \SRef{AMEvolv}.
         \Intro{\AME}{application model elements}
         
   \item Time, essential to evolution, is incorporated into the theory
         through the algebraic structure 
         \Intro{\TimeAxis}{time axis} $\TimeAxis=\tuple{\Time,F}$, 
         where $\Time$ \Intro{\TimeAxis}{time axis} is a (discrete,
         totally ordered) time axis,
         and $F$ a set of functions over $\Time$.
         For the moment, $F$ is assumed to contain the one-step increment
         operator $\Inc$, and the comparison operator $\leq$.
         Several ways of defining a time axis exist, see e.g.
         \cite{Article:87:Clifford:Time}, \cite{Article:91:Wiederhold:TimeGranularity}
         or \cite{Article:84:Allen:ActionTime}.

         The time axis is the axis along which the application model evolves.
         With this time axis, an \keyword{application model history}
         is a (partial) mapping $\Time \PartFunc \AME$.%
         \footnote{In this article, $\PartFunc$ is used for partial functions,
         and $\Func$ for total functions.}
         $\AMH$ is the set of all such histories.
         In a later section, we will pose well-formedness restrictions on
         histories.

         Other time models are possible, for example, in distributed systems
         a relative time model might be used.
         For a general survey on time models,
         see \cite{Article:92:Roddick:TimeSurvey}.
         The linear time model is usually chosen in historical databases
         (see for example \cite{Article:90:Snodgrass:TempDBStatus}).

   \item $\Tasks$ is the domain for actions that can be performed on
         application model histories.
         \Intro{\TaskDef}{action specifications}
         
   \item The semantics of the actions in $\Tasks$ is provided by  
         the state transition relation on application model histories: 
         $\StateTrans{~}{~}
           \subseteq \Tasks \Carth \Time \Carth \AMH \Carth \AMH$,
         where $H \StateTrans{m}{t} H'$ means: $H'$ may result after
         applying action $m$ to $H$ at time $t$.
         In business applications,  most actions will turn out to be
         deterministic. However, sometimes it is useful to allow for
         nondeterminism; for example when external influences can
         effect the outcome of a process, while these influences
         themselves are not considered part of the universe of
         discourse.

         Our way of abstracting the semantics of actions was inspired
         by the Temporal Logic of Actions as discussed in [Lam91].

         \Comment{
         The semantics of an action $m \in \Tasks$ are assumed to be defined
         by means of a suitable (temporal) logic. 
         For instance, TLA (Temporal Logic of Actions) as defined in
         \cite{Report:91:Lamport:TLA}.
         }
         \Intro{\StateTrans{m}{t}}{action semantics}
\end{enumerate}
In the example of \SRef{RunningExample}, the application model history
is told by the evolution of its elements,
such as \LISA{LP}, \LISA{CD} and \LISA{Recording}.
Their histories are summarized in \SRef{\ExHistory}.
\TablE[\ExHistory]{Some element histories}
%

%% file: hu0206.tex
\subsection{A dual vision}
\SLabel{subsection}{DualVision}

The execution of an action at some point of time is referred
to as an \Keyword{event occurrence}.
The domain of sequences of event occurrences is identified by:
\begin{definition}(event occurrence sequence domain)
   \[ \EvOcc = \Time \PartFunc \Tasks \]
   \Intro{\EvOcc}{event occurrence sequences}
\end{definition}

An application model history ($H$) describes the evolution of
an underlying application.
A prefix of this history describes the evolution of this application
upto some point of time,
and forms a \Keyword{state}[evolving information system state]
of an associated evolving information system.
First we introduce prefixing of a single element evolution:
\begin{definition}(element evolution prefix)
   If $h :\Time \PartFunc \AME$ then the prefix of $h$ at time $t$
   is:
   \[ \UpTo{h}{t} = \LambdaFun{s}{\SIf{s \leq t}{h(s)}{h(t)}} \]
   \Intro{\UpTo{h}{t}}{history prefixing}
\end{definition}
Some obvious properties, concerning idempotence, for history prefixing are:
\begin{lemma}[SingleLaterIdempotence]
   If $h :\Time \PartFunc \AME$, and $u \leq t$ then:
   \[ \UpTo{(\UpTo{h}{t})}{u}=\UpTo{h}{u} \]
\end{lemma}
\Comment{
\begin{proof}
   Let $u \leq t$, then:
 
   $\UpTo{(\UpTo{h}{t})}{u}$ \\
   $\EqTo \Explain{definition of $\UpTo{(\UpTo{h}{t})}{u}$}$
 
   $\LambdaFun{s}{\SIf{s \leq u}{\UpTo{h}{t}(s)}{\UpTo{h}{t}(u)}}$\\
   $\EqTo \Explain{definition of $\UpTo{h}{t}$}$

   $\LambdaFun{s}{}\!\!\!\!
       \begin{array}[t]{l}
          ~\SIf{s \leq u}{\\
          \Tab \LambdaFun{s}{\SIf{s \leq t}{h(s)}{h(t)}}\!\!(s) \\
          }{\\ 
          \Tab \LambdaFun{s}{\SIf{s \leq t}{h(s)}{h(t)}}\!\!(u) \\
          }
       \end{array}$\\
   $\EqTo \Explain{assumption: $u \leq t$}$

   $\LambdaFun{s}{\SIf{s \leq u}{h(s)}{h(u)}}$\\
   $\EqTo \Explain{definition of $\UpTo{h}{u}$}$

   $\UpTo{h}{u}$
\end{proof}
}
\begin{lemma}[SingleEarlierIdempotence]
   If $h :\Time \PartFunc \AME$, and $t \leq u$ then:
   \[ \UpTo{(\UpTo{h}{t})}{u}=\UpTo{h}{t} \]
\end{lemma}
\Comment{
\begin{proof}
   Let $t \leq u$, then:
 
   $\UpTo{(\UpTo{h}{t})}{u}$ \\
   $\EqTo \Explain{definition of $\UpTo{(\UpTo{h}{t})}{u}$}$
 
   $\LambdaFun{s}{\SIf{s \leq u}{\UpTo{h}{t}(s)}{\UpTo{h}{t}(u)}}$\\
   $\EqTo \Explain{definition of $\UpTo{h}{t}$}$

   $\LambdaFun{s}{}\!\!\!\!
       \begin{array}[t]{l}
          ~\SIf{s \leq u}{\\
          \Tab \LambdaFun{s}{\SIf{s \leq t}{h(s)}{h(t)}}\!\!(s) \\
          }{\\ 
          \Tab \LambdaFun{s}{\SIf{s \leq t}{h(s)}{h(t)}}\!\!(u) \\
          }
       \end{array}$\\
   $\EqTo \Explain{assumption: $t \leq u$}$

   $\LambdaFun{s}{\SIf{s \leq u}{h(s)}{h(t)}}$\\
   $\EqTo \Explain{definition of $\UpTo{h}{t}$}$

   $\UpTo{h}{t}$
\end{proof}  
}
The states of an evolving information system,
tracking application model history $H$,
are identified by:
\begin{definition}(evolving information system state)
   If $H \in \AMH$ then the state of $H$ at time $t$
   is:
   \[ \UpTo{H}{t} = \Set{\UpTo{h}{t}}{h \in H} \]
\end{definition}
Note that each state of an evolving information system is 
an application model history as well ($\UpTo{H}{t} \in \AMH$).
States are also referred to as \Keyword{initial histories}[initial history].
The result of \SRef{SingleLaterIdempotence} and \SRef{SingleEarlierIdempotence}
can be generalised to:
\begin{corollary}[PrefixIdempotence]
   If $H$ is an application model history, then
   \begin{eqnarray*}
      u \leq t & \implies & \UpTo{(\UpTo{H}{t})}{u}=\UpTo{H}{u} \\
      t \leq u & \implies & \UpTo{(\UpTo{H}{t})}{u}=\UpTo{H}{t}
   \end{eqnarray*}
\end{corollary}

The evolution of an application model is described by an application model
history $H$.
Besides, this evolution may be modelled as a
sequence $E$ of event occurrences,
specifying subsequent changes to initial histories of the application model,
starting from the initial application model.
Thus the combination of $E$ and $H$ leads to a dual vision on states
of evolving information systems.
On the one hand, a state results from a set of event occurrences.
On the other hand,
a state is a prefix of an application model history.

In \SRef{\BehavesRelation}, a broader view on this duality is depicted.
A user observes the history of a universe of discourse ($H_{uod}$), and
formulates the observed evolution by means of a set of event occurrences
($E$).
These event occurrences result, by means of the $\Behaves$ relation, in an
application model history ($H_{is}$) as stored in the information system.
This latter application model history must be validated (emperically) against
the observed history of the universe of discourse.
\Fig[\BehavesRelation]{A user observing the history of a universe of discourse}

\Comment{
Hierbij de volgende meta-opmerking maken:
remarks zijn dingen die leuk zijn om te lezen, maar ze horen
niet bij de hoofdlijn van het betoog.
\begin{remark}
   The relation between $H$ and $E$ can be intuitively formulated as:
   \[ {d \UpTo{H}{t} \over d t}
      =
      \left\{ \begin{array}{lcl}
        E(t)      & ~~~ & \mbox{if $\DefinedAt{E}{t}$} \\
        \emptyset & ~~~ &  \mbox{otherwise}
      \end{array} \right.
   \]
\end{remark}
}
The relation between an application model history $H$, and a set of
event occurences $E$ is captured by the $\Behaves$ predicate:
\begin{definition}
   Let $E \subseteq \EvOcc$ and $H \in \AMH$, then:
   \begin{eqnarray*}
      \lefteqn{\Behaves(E,H)}\\
         & \Eq   & 
         \Al{\tuple{t,m} \in E
           }{\UpTo{H}{t} \StateTrans{m}{t} \UpTo{H}{\Inc t} \Forc{3}
           } \\
         & \land &
         \Al{t}{\UpTo{H}{t} \not= \UpTo{H}{\Inc t} \implies
                  \Ex{m}{\tuple{t,m} \in E} \Forc{3}}
   \end{eqnarray*}
   \Intro{\Behaves}{well-behaved information system}
\end{definition}
The first part of the above definition states that every event occurence
must be reflected in the application model history $H$.
On the other hand, the second part of the definition states
that any change in the $H$ must be based on some event occurence.

The events which are described in our running example are:
\begin{enumerate}
\item event $E_1$ occurring at time $t_1$: the introduction of CD's
\item event $E_2$ occurring at time $t_2$: the abolishment of LP's
\end{enumerate}
For simplicity, we assume that no other events (including changes
to the population) have taken place.
If we refer to application model history of this example by
the name $Store$, then the following three
different states can be recognized:
\begin{enumerate}
\item $\UpTo{Store}{t_1}$: the initial history of the system,
      until CD's are introduced.
\item $\UpTo{Store}{t_2}$: the history of the system after the
      introduction of CD's,
      upto the abolishment of LP's (at time $t_2$).
\item $\UpTo{Store}{t_3} = \UpTo{Store}{\Inc t_2}$
      for points of time later than $\Inc t_2$.
\end{enumerate}
Then the predicate $\Behaves$ enforces the following properties:
\begin{enumerate}
\item $\UpTo{Store}{t_1} \StateTrans{E_1}{t_1} \UpTo{Store}{t_2}$
\item $\UpTo{Store}{t_2} \StateTrans{E_2}{t_2} \UpTo{Store}{t_3}$
\end{enumerate}
Due to this property,
the communication between user and information system can be transaction
oriented.
The description of a (convenient) language for this communication
falls outside the scope of this paper.
For more details see
\cite{Article:89:Kim:Orion}
\cite{Article:89:Bretl:Gemstone}
\cite{Article:92:Tresch:OOEvolution}
\cite{PhdThesis:94:Proper:EvolvConcModels},
and \cite{Report:93:Proper:DisclSch}.
\Comment{
\begin{remark}
States of an evolving information system are finite approximations of the
associated aplication model history, or:
\[ \lim_{t \into \infty} \tuple{\UpTo{H}{t},\UpTo{E}{t}}
  =
  \tuple{H,E}
\]
\end{remark}
}

At this point,
we have demarcated the states and transitions of an evolving information
system.
Later,
we will impose wellformedness restrictions on application model histories,
and thus on the states of the evolving information system.
We will use $\IsAMH(H)$ to denote that $H$ satisfies these restrictions.
These restrictions on states imply a restriction on transitions,
expressed by the predicate $\IsEIS$:
\begin{definition}
   Let $E \in \EvOcc$ and $H \in \AMH$, then:
   \begin{eqnarray*}
      \IsEIS(E,H) & \Eq   & \Behaves(E,H) \\
                  & \land & \Al{t \in \Time}{\IsAMH(H_t)}
   \end{eqnarray*}
   \Intro{\IsEIS}{well-behaved proper information system}
\end{definition}

\Comment{
\begin{remark}
   In the course of the life time of an evolving information system, the set
   of event occurrences will grow. 
   From the above definition, we see that $\UpTo{H}{t}$ grows monotoneously
   in the course of time if $\UpTo{E}{t}$ grows.
   This can be compared to the notion of entropy in thermo dynamics.
   Entropy is a physical measure for chaos, and can not decrease in time
   (see for example \cite{Article:90:Hawking:TimeArrow}).
\end{remark}
}

%% file: eu0207.tex
\subsection{Dissecting application model histories}

The current version (snapshot) of an application model is constituted
by the current versions of all application model elements.
This allows us to reconstruct the sequence of successive versions of the
application model.
This reconstruction will be described in \SRef{AMEvolv}.

An application model version contains, as its two main components, an
information structure version and an action model version (see 
\SRef{\ModelsHierarchy}).
The information structure and action model are assumed to be described in
some modelling technique.
The only assumption we make on this modelling technique, is that it spans
an application model universe, and an information structure universe in
particular.
These universes provide the state space for the evolution of
application models and information structures.
In the following two sections, the information structure universe and
application model universe are defined respectively. 
As stated before,
in a later section some well-formedness rules are given, limiting the
freedom of changes that can be performed upon application model versions 
in the course of time.

%% file: hu03.tex
\section{Generalised Information Structures}
\SLabel{section}{AppMod}

The kernel of the application model universe is formed by the 
\Keyword{information structure universe},
fixing the evolution space for information structures.
Therefore, we take this universe as a starting point to build
the formal framework,
as it forms a solid (time and application independent)
base for this framework.

\input{hu0301} 
\input{gu0302} 

%% file: hu0301.tex
\subsection{The information structure universe}

The information structure universe, for a given modelling technique, is defined as:
\begin{definition}
   The universe $\InfStructUniverse$ for information
   structures is determined by the structure:
   \[
      \InfStructUniverse = 
      \tuple{\LTypes,\NLTypes,\TypeRel,\ParentOf,\IsSch}
   \]
\end{definition}
where $\LTypes$ are label object types, $\NLTypes$ are abstract object
types.
The relation $\TypeRel$ captures relatedness between object
types.
Inheritance of identification of object types is described in the
relation $\ParentOf$.
Finally, the predicate $\IsSch$ (is schema) embodies wellformedness of information
structures.
These components are discussed in more detail in the next subsections.

Further refinements of the information structure universe depend
on the chosen data modelling technique (such as NIAM, ER, PSM and Object 
Oriented data models), and are beyond the scope of the theory.
An important refinement of any concrete data modelling technique will
be the recognition of relationship types between object types.
For instance, the fact that in figure 5 Recording is a relationship type
between CD and Song, is of no importance to the general theory.
Abstracting from such relationships between object types was essential
because this is exactly where most modelling techniques (in particular OO
from ER or NIAM) differ the most.

In \SRef{ExER} we will see how ER fits within this framework.
For more examples, see \cite{PhdThesis:94:Proper:EvolvConcModels}
and \cite{Report:93:Proper:EvolvPSM}.
For our purposes, an information structure universe is assumed to provide (at
least) the above components, which are available in all conventional high
level data modelling techniques.

\subsubsection{Object types}
The central part of an information structure is formed by its object types
(referred to as object classes in object oriented approaches).
Two major classes of object types are distinguished. 
Object types who's instances can be represented directly (denoted) on a medium
(strings, natural numbers, etc)
form the class of label types $\LTypes$ \Intro{\LTypes}{label types}.
The other object types,
for instance entity types or fact (relation) types,
form the class \Intro{\NLTypes}{non label types} $\NLTypes$.
The set of all possible object types is defined as:
$\OTypes = \LTypes \union \NLTypes$. \Intro{\OTypes}{object types}
The example of \SRef{\LPLibrary} contains nine object types: 
three entity types \LISA{Record}, \LISA{Song} and \LISA{Frequency},  
two relation types \LISA{Recording} and \LISA{Lending-frequency}, 
and four label types \LISA{Title}, \LISA{Artist}, \LISA{Author} and 
\LISA{Times}. 

\subsubsection{Type relatedness}
The relation $\TypeRel ~\subseteq \OTypes \Cart \OTypes$ expresses 
\Keyword{type relatedness} between object types 
(see \cite{Report:91:Hofstede:PSM}).
\Intro{\TypeRel}{type relatedness}

Object types $x$ and $y$ are termed type related 
($x \TypeRel y$) iff populations of object types $x$ and $y$ 
{\em may} have values in common in any version of the application model.
Type relatedness corresponds to mode equivalence in programming languages
(\cite{Book:76:Wijngaarden:Algol68}).
The relation of type relatedness can be recognised in conventional 
modelling techniques like ER, NIAM, or PSM, as well as in semantic
data model approaches including object oriented concepts
(see for example \cite{Article:85:Cardelli:DBTypeTheory}).
Typically, subtyping and generalisation lead to type related object types. 
For the data model depicted in \SRef{\LPLibrary}, the type 
relatedness relation is the identity relation: $x \TypeRel x$ for all 
object types $x$.

An example of a more complex type relatedness relation is provided in the
PSM data model in \SRef{\RootCS}.
In this example,
the administration of a broker for both boats and houses is modelled.
The solid arrow stands for a subtyping (specialisation) relation,
whereas the dotted arrows represent generalisations.
A major difference between generalisation and specialisation is that
the population of subtypes is defined by means of a subtype defining
rule in terms of the population of the supertype,
whereas a generalised object type directly inherits the complete populations
from its specificers
(\cite{Article:87:Abiteboul:IFO},\cite{Report:91:Hofstede:PSM}).
The type relatedness relation for the data model  of \SRef{\RootCS}
is therefore:
\begin{enumerate}
\item $\LISA{Product} \TypeRel \LISA{Boat}$
\item $\LISA{Product} \TypeRel \LISA{House}$
\item $\LISA{Product} \TypeRel \LISA{Real estate}$
\item $\LISA{Real estate} \TypeRel \LISA{Boat}$
\item $\LISA{Real estate} \TypeRel \LISA{House}$
\end{enumerate}
\FiG[\RootCS]{A data model with generalisation and specialisation}

According to the the intuitive meaning of type relatedness,
this relation is required to be reflexive and symmetrical:
\begin{Axiom}{ISU}[ReflexTypeRel](reflexive)
   $x \TypeRel x$
\end{Axiom}
\begin{Axiom}{ISU}[SymTypeRel](symmetrical)
   $x \TypeRel y \implies y \TypeRel x$
\end{Axiom}

\subsubsection{The identification hierarchy}
In data modelling, a crucial role is played by the notion of object 
identification:
each object type of an information structure should be identifiable.
In a subtype hierarchy however,
a subtype inherits its identification from its
super type, whereas in a generalisation hierarchy the identification of
a generalised object type is inherited from its specifiers. 
For the data model depicted in \SRef{\RootCS} this means that
instances of \LISA{Real estate} are identified in the same way as instances of
\LISA{Product}.
The identification of instances from \LISA{Product} depends on the
identification of instances from \LISA{Boat} or \LISA{House}
(note that an instance from \LISA{Product} is either an
instance from \LISA{Boat} or an instance from \LISA{House}).
For the data model depicted in \SRef{\LPCDLibrary}, it means
that instances of \LISA{LP} and \LISA{CD} are identified in the same way as 
instances of \LISA{Medium}.

An object type from which the identification is inherited, is termed an 
\Keyword{ancestor} of that object type.
The inheritance hierarchy (identification hierarchy)
is provided by the relation $x \ParentOf y$, 
meaning $x$ is an ancestor of $y$.
\Intro{\ParentOf}{ancestor relationship}
For \SRef{\LPCDLibrary} this leads to: $\LISA{Medium} \ParentOf \LISA{LP}$
and $\LISA{Medium} \ParentOf \LISA{CD}$.
The inheritance relation is both transitive and irreflexive.
\begin{Axiom}{ISU}[TransParentOf](transitive)
   $x \ParentOf y \land y \ParentOf z \implies x \ParentOf z$
\end{Axiom}
\begin{Axiom}{ISU}[ReflexParentOf](irreflexive)
   $\neg x \ParentOf x$
\end{Axiom}
Similar axioms can be found as properties in literature about typing 
theory for databases
(\cite{Article:90:Bruce:DBTypeTheory},
\cite{Article:90:Ohori:DBTypeTheory} and
\cite{Article:85:Cardelli:DBTypeTheory}).
The difference, between these properties and ours, lies in the 
abstraction of an underlying structure
of object types and their instances.
As we do not make any assumption on these structures,
such properties must be stated as axioms.
Another reason is that the
inheritance hierarchy is intertwined with type relatedness, requiring
appropriate axioms.

Object types without an ancestor, are called \Keyword{roots}[root]:
  \Intro{\IsRoot}{root object type} 
  $\IsRoot(x) \Eq \neg\Ex{z}{z \ParentOf x}$.
The roots $x$ of an object type $y$ are found by:
  \Intro{\RootOf}{root of object type}
  \[ x \RootOf y \Eq (x=y \lor x \ParentOf y) \land \IsRoot(x) \]
The finite depth of the inheritance hierarchy is expressed by the 
following schema of induction:
\begin{Axiom}{ISU}[ParentIndSch](ancestor induction)
   If $\forall_{x \ParentOf y}[F(x)] \implies F(y)$ for any $y$,
   then $\forall_{x}[F(x)]$.
\end{Axiom} 

\Comment{
Note that this schema of induction does not have an explicit {\em base step}.
One might expect the induction schema to be:
\begin{quote}
   If $\IsRoot(y) \implies F(y)$ (\Keyword{base step})\\ 
   and  $\forall_{x \ParentOf y}[F(x)] \implies F(y)$ (\Keyword{induction step})\\
   for all $y$, then $\forall_{x}[F(x)]$.
\end{quote}
However, the $\IsRoot(y)$ case is (see the definition of $\IsRoot$) already 
included in $\forall_{x \ParentOf y}[F(x)] \implies F(y)$. 
} 

From the intuition behind the ancestor relation it follows that object types
may have instances in common with their ancestors.
This implies that object types not only inherit identification from their
ancestors, but type relatedness as well.
These requirements are laid down in the following axioms:
\begin{Axiom}{ISU}[TypeRelInheritance](inheritance of type relatedness)
   \[ x \TypeRel y \land y \ParentOf z \implies x \TypeRel z \]
\end{Axiom}
\begin{Axiom}{ISU}[FoundedTypeRel](foundation of type relatedness)
   \[ x \TypeRel y \land \neg \IsRoot(y) \implies
      \exists_z[x \TypeRel z \land z \ParentOf y] \]
\end{Axiom}

For every data model from conventional data modelling techniques, an
ancestor and root relation can be derived. 
If no specialisations or generalisations are present in a particular 
data model, the associated ancestor relation will be empty. 
As a result, the root relation will then be the identity relation.
For instance the root relation for \SRef{\LPLibrary} is:
$x \RootOf x$ for every object type $x$.
When the data model at hand contains specialisation or generalisations,
the relations $\ParentOf$ and $\RootOf$ will be less trivial.
In \SRef{\RootDepGr}, the relation $\RootOf$
of the data model in \SRef{\RootCS} 
is provided.
This graph is a so called {\it root dependency graph}, as it
depicts the dependency of object types on their roots.
\Fig[\RootDepGr]{Root dependency graph}

\subsubsection{Correctness of information structures}
An information structure is spanned by a set of object types.
Not all sets of object types taken from $\OTypes$ will correspond
to a correct information structure.
Therefore, a technique dependent predicate 
$\IsSch \subseteq \Powerset(\OTypes)$ has to be supplied,
designating which sets of object types form a correct information
structure.
\Intro{\IsSch}{correct information structure}

\subsubsection{An example: ER}
\SLabel{subsubsection}{ExER}
As a brief example of how the general theory can be related to an existing
modelling technique, we consider ER in this section.
As stated before, a fully elaborated and formalised application of the
theory to an object-role modelling technique can be found in
\cite{Report:93:Proper:EvolvPSM}.

For Chen's (\cite{Article:76:Chen:ER}) ER model (extended with subtyping), 
the information structure universe will be:
\begin{description}
   \item[Label Types] The set of label types $\LTypes$ in ER corresponds
	to the printable attribute types.
	Note that in some ER versions, entity types can be used as
	attribute for other entity types.

   \item[Non-Label Types] The set of non-label types $\NLTypes$ is defined
	as the set of relationship types, entity types and associative object
	(entity) types.
	
   \item[Inheritance] Traditional ER only contains the notion of subtyping.
	So for each subtype $x$ of a supertype $y$ we have: $y \ParentOf x$.
	The complete inheritance relation $\ParentOf$ is then obtained by
	applying the transitive closure.
        
   \item[Type Relatedness] Two subtypes of the same supertype are type
        related.
        Furthermore, subtyping is the only way in ER to make type related 
        object types.
        Furthermore, a subtyping hierarchy has a unique top element.
	Let $\Top(x)$ denote the unique top element of the subtyping
	hierarchy containing object type $x$.
        As a result, type relatedness for ER is defined as:
	$x \TypeRel y \Eq \Top(x)=\Top(y)$. 

   \item[Schema Wellformedness] The predicate $\IsSch$ can be described
        according to ER rules.
        This will be omitted in this paper.
\end{description}
The information structure universe axioms are easily verified.
The type relatedness axioms ISU1 and ISU2 are immediate consequences
of the above definition.
The identification hierarchy axioms ISU3, ISU4 and ISU5
directly follow from the nature of subtyping in ER.
The axioms that relate type relatedness with the identification hierarchy
are also easily verified.

%% file: gu0302.tex
\subsection{Properties of information structure universes}

The axioms so far try to model the concepts of type relatedness, object type and
inheritance.
In this section, we derive some usefull properties of information structure 
universes, illustrating the validity of the \ClassRef{ISU} axioms at the same 
time.
The first property relates the root relationship to type relatedness:
\begin{lemma}
   Any root of an object type is related to that object type:
   \[ x \RootOf y \implies x \TypeRel y\]
\end{lemma}
\begin{proof}%
  $x \RootOf y$\\
  $\Therefore$ \Explain{definition of $\RootOf$}%
  \hspace{-0.7em}%
  \footnote{In proofs, the $\!\!\!\!\!\Therefore$ symbol stands for a logical deduction step.
            A logical equivalence in a proof is denoted by $\!\!\!\!\EqTo$.
            The motivation for the deduction step is provided in
            parenthesis.}
 
  $x \ParentOf y$\\
  $\Therefore$ \Explain{definition of $\TypeRel$}

  $x \TypeRel x \land x \ParentOf y$\\
  $\Therefore$ \Explain{\SRef{TypeRelInheritance}}

  $x \TypeRel y$
\end{proof}

Axiom~\ref{FoundedTypeRel} may be generalized to:
\begin{lemma}(common roots)[CommonRoots]
   Sharing a root is equivalent with being type related:
   \[ x \TypeRel y \iff \exists_z[x \TypeRel z \land z \RootOf y] \]
\end{lemma}
In order to prove this property, and interesting properties to come, two
proof schemas concerning {\em inheritance} and {\em foundation} of
properties are introduced first.
We call a property $P$ of object types a
\Keyword{strong inheritance property},
iff for all $x,y$: \[ P(x) \land x \ParentOf y \implies P(y) \]
Note that \SRef{TypeRelInheritance} states that the relation $P_x$,
defined by $P_x(y) = x \TypeRel y$,
is a strong inheritance property for all $x$.
A property $P$ will be referred to as a
\Keyword{weak inheritance property}
iff, for all $y$:
\[ P(y) \land \neg \IsRoot(y) \implies \exists_x[P(x) \land x \ParentOf y] \]
Axiom~\ref{FoundedTypeRel} states that the relation $P_x$,
defined by $P_x(y) = x \TypeRel y$,
is a weak inheritance property for all $x$.
The first proof schema is rather straightforward, and is concerned with
inheritance of properties:
\begin{theorem}(inheritance schema)[InheritanceSchema]
   If $P$ is a strong inheritance property, then the property
   is preserved by the $\RootOf$ relation:
   \[ P(x) \land x \RootOf y \implies P(y) \]
\end{theorem}
\begin{proof}
   Suppose $P$ is a strong inheritance property, then:

   $P(x) \land x \RootOf y$\\
   $\Therefore$ \Explain{definition of $\RootOf$}

   $P(x) \land (x \ParentOf y \lor x=y)$\\
   $\Therefore$ \Explain{$P$ is a strong inheritance property}

   $P(y)$
\end{proof}
The second proof schema is concerned with the foundation of properties:
\begin{theorem}(foundation schema)[FoundationSchema]
   If $P$ is a weak inheritance property, then $P$ 
   originates from root object types.
   \[ P(y) \implies \exists_x[P(x) \land x \RootOf y] \]
\end{theorem}
\begin{proof}
   Suppose $P$ is a weak inheritance property.
   Now we apply ancestor induction in order to prove the 
   theorem.
   From the induction hypothesis it follows:
   \[ \forall_{v \ParentOf w}[
         P(v) \implies \exists_z[P(z) \land z \RootOf v]
      ] 
   \]
   From this we will prove:
   \[ P(w) \implies \exists_z[P(z) \land z \RootOf w] \]

   Suppose $P(w)$. 
   The $\IsRoot(w)$ case is obvious, due to the reflexivity of the 
   $\RootOf$ relation.
   For $\neg \IsRoot(w)$ we get:

   $P(w) \land \neg \IsRoot(w)$\\
   $\Therefore$ \Explain{$P$ is a weak inheritance property}

   $\exists_u[u \ParentOf w \land P(u)]$\\
   $\Therefore$ \Explain{induction hypothesis} 
   
   $\exists_u[( P(u) \implies \exists_z[P(z) \land z \RootOf u] )
              \land u \ParentOf w \land P(u)]$\\ 
   $\Therefore$ \Explain{elaborate} 

   $\exists_u[\exists_z[P(z) \land z \RootOf u] \land
              u \ParentOf w]$ \\ 
   $\Therefore$ \Explain{definition of $\RootOf$}

   $\exists_z[P(z) \land z \RootOf w]$
\end{proof}
With these proof schemata we get an elegant proof of \SRef{CommonRoots}:
\begin{proof}
   We prove both directions separately:
   \RtoL

   Define $P(a)=x \TypeRel a$. 
   We will apply this to \SRef{TypeRelInheritance}:
   
   $P(y) \land y \ParentOf z \implies P(z)$\\
   $\Therefore$ \Explain{$P$ is a strong inheritance property}

   $P(y) \land y \RootOf   z \implies P(z)$\\
   $\Therefore$ \Explain{elaborate}

   $\exists_y[P(y) \land y \RootOf z] \implies P(z)$\\
   $\Therefore$ \Explain{definition of $P$}
 
   $\exists_y[x \TypeRel y \land y \RootOf z] \implies x \TypeRel z$

   \LtoR

   Define $P(a)=x \TypeRel a$. 
   We will apply this to \SRef{FoundedTypeRel}.

   $P(y) \land \neg \IsRoot(z) \implies \exists_z[P(z) \land z \ParentOf y]$\\
   $\Therefore$ \Explain{$P$ is a weak inheritance property}

   $P(y) \implies \exists_z[P(z) \land z \RootOf y]$\\
   $\Therefore$ \Explain{definition of $P$}

   $x \TypeRel y \implies \exists_z[x \TypeRel z \land z \RootOf y]$
\end{proof}
The result of the above lemma can be generalised to the
following theorem:
\begin{theorem}(type relatedness propagation)[TypeRelPropagation]
   Type relatedness of roots is equivalent with type relatedness of object 
   types:
   \[ \exists_{z_1 \TypeRel z_2}[
         z_1 \RootOf x \land z_2 \RootOf y] \iff
      x \TypeRel y \]
\end{theorem}
\begin{proof}\\
   $\exists_{z_1,z_2}[
       z_1 \TypeRel z_2 \land z_1 \RootOf x \land z_2 \RootOf y]$\\
   $\EqTo$ \Explain{apply \PSRef{CommonRoots} to $z_1$}

   $\exists_{z_2}[x \TypeRel z_2 \land z_2 \RootOf y]$\\ 
   $\EqTo$ \Explain{apply \PSRef{CommonRoots} to $z_2$}

   $x \TypeRel y$
\end{proof}

\Fig[\TypeRelPropag]{Data model with propagation of type relatedness}
As an illustration of this theorem, consider the PSM data model from
\SRef{\TypeRelPropag}.
It contains two generalisations, two specialisations, and two power types
($D, E$).
Power types are the data modelling pendant of powersets used in set theory.
The instances of object types $D$ and $E$ are sets of instances of $B$ and
$C$ respectively.
The $\RootOf$ relation for this data model, is given in \SRef{\TypeRelDepGr}.
The type relatedness of $D$ and $E$, which itself follows from the type
relatedness of $B$ and $C$ (\cite{Report:91:Hofstede:PSM}),
is propagated to $F$ and $G$ by means of the $\RootOf$ relationship
and \SRef{TypeRelPropagation}.
In \cite{Report:91:Hofstede:PSM}, \cite{Report:91:Hofstede:LISA-D}, 
the inheritance of type relatedness via type constructions, e.g. 
powertyping, is elaborated.
\Fig[\TypeRelDepGr]{Root dependency graph showing propagation of type
                    relatedness}

%% file: gu04.tex
\section{Generalised Application Models}

Besides the information structure, an application model contains a
number of other elements.
The hierarchy of models in \SRef{\ModelsHierarchy} describes
how an application model is constructed from other (sub)models.
However,
this hierarchy disregards relations that must hold between the submodels,
for example, how a population relates to the information structure.
These relations are crucial elements of an application model,
as they form the fabric of the application model.

An \keyword{application model version} provides a complete description of
the state of the information system at some point of time.
Such an application model version is bound to the 
\Keyword{application model universe} $\AppModUniverse$.
\begin{definition}
   An application model universe is spanned by the tuple:
   \Intro{\AppModUniverse}{application model universe}
   \[
      \AppModUniverse =
         \tuple{\InfStructUniverse, 
                \ConcrDoms, \UniDom, \IsPop, 
                \ConstrDef, \TaskDef, \StateTrans{~},
                \Depends}
   \]
\end{definition}
where the information structure universe $\InfStructUniverse$ has been
introduced in the previous section.
$\ConcrDoms$ is a set of underlying concrete domains to be associated to
label types.
The set $\UniDom$ is derived from these concrete values,
and is a domain for instantiating abstract object types.
The predicate $\IsPop$ checks if such an instantiation
is wellformed.
$\ConstrDef$ and $\TaskDef$ are the universes for constraint and method
definitions respectively.
The semantics of both constraints and methods is provided by
the ternary predicate $\StateTrans{~}{~}$ (see \SRef{EIS}).
The dependencies of constraints and method on the type level
($\OTypes$, $\LTypes \Carth \ConcrDoms$) are described by the relation
$\Depends$.
The information structure universe $\InfStructUniverse$ was introduced in the
previous section.
The other components of the application model universe are discussed in the
remainder of this subsection.

\subsection{Domains}
The separation between concrete and abstract world is
provided by the distinction between the information structure
$\InfStruct$, and the set of underlying (concrete) domains in
$\ConcrDoms$ (\cite{Report:91:Hofstede:LISA-D}).
\Intro{\ConcrDoms}{concrete domains} 
Therefore, label types in an information structure version will have 
to be related to domains.
An application model version contains a mapping $\Domain_t$
providing the relation between label types and domains.
Each domain assignment $\Domain_t$ is bound to:
\Intro{\Domain}{domain assignments}
\[ \Domain = \LTypes \PartFunc \ConcrDoms \]
Some illustrative examples of such domain assignments, in the context of
the rental store running example, are: 
\LISA{%
   Times $\mapsto$ Natno,
   Title $\mapsto$ String,
}
where \LISA{Natno} and \LISA{String} are assumed to be (names of) concrete
domains.

\subsection{Instances}
The population of an information structure is not, as usual, a partial 
function that maps object types to sets of instances. 
Rather, an instance is considered to be an independent thing, which can evolve
by itself.
Therefore, (non empty) sets of object types are associated to instances, 
specifying the object types having this instance as an instantiation.
This association is the intuition behind the relation $\HasTypes_t$.
The domain for this relation is:
\Intro{\HasTypes}{types instance relation}
\[ 
   \HasTypes = \UniDom \Carth (\Powerset(\OTypes) \SetMinus \set{\emptyset}) 
\]
where $\UniDom$ is the set of all possible instantiations of object types.
\Intro{\UniDom}{all possible instantiations}
Note that $\HasTypes_t$ is a relation rather than a (partial) function.
The reason is to support complex generalisation hierarchies.
For example, suppose that $\set{a_1,a_2}$ is an instance of both
$D$ and $E$ in \SRef{\TypeRelPropag}.
Then $\set{a_1,a_2} \HasTypes_t \set{D,F}$ and
 $\set{a_1,a_2} \HasTypes_t \set{E,G}$.

Another example of such an association is
$\tuple{l_1,\set{\LISA{Medium, Lp}}}$, meaning $l_1$ is an (abstract) 
instance of entity types \LISA{Medium} and \LISA{Lp}.
The population of an object type, traditionally provided as a function
$\Pop: \OTypes \Func \Powerset(\UniDom)$, can be derived from the
association between instances and object types:
$\Pop_t(x) = \Set{v}{v \HasTypes_t Y \land x \in Y}$.
\Intro{\Pop}{population of object types}

Not all subsets of $\HasTypes$ will correspond to a proper population.
A population of an information structure will have to adhere to some
technique dependent properties.
These properties are assumed to be provided by the predicate
$\IsPop \subseteq \Powerset(\OTypes) \Carth \Powerset(\HasTypes)$.
Note that this predicate does not take the validity of constraints in
the application model into consideration.
This is not yet possible, as constraints may be transition oriented,
implying that they can only be enforced in the context of the evolution of 
the elements.
The enforcing of constraints on the (evolution of) populations will therefore
be postponed until \SRef{AMEvolv}.

\subsection{Constraints}
Most data modelling techniques offer a language for expressing constraints,
both state and transition oriented.
This language describes a set $\ConstrDef$ of all possible constraint
definitions.

Each constraints $C$ is treated as a partial function,
assigning constraint definitions to object types:
$C: \OTypes \PartFunc \ConstrDef$.
Constraint $C$ is said to be {\em owned} by object type $x$,
if $x$ has assigned a constraint definition by constraint $C$.
Each constraint is considered to be an application model element.

Constraints are inherited via the identification hierarchy.
However, as in object oriented data modelling techniques,
overriding (strengthening) of constraint definition in identification
hierarchies is possible
(see for instance \cite{Article:91:Troyer:CAISE91}).
This will be discussed in more detail in a later section
as \SRef{AMVStrength}.

\FiG[\Airplanes]{Constraint assignment}
As an illustration of the assignment of constraints to object types,
consider \SRef{\Airplanes}. 
The depicted data model is conforming to NIAM, while the subtype 
defining rules have been formulated in LISA-D.
The modelled universe of discourse is concerned with the administration of 
airplanes.
As airplanes should be replaced in time,
the age of an airplane is an important attribute.
Furthermore, an airplane may be registered by an aviation association,
in which case it has associated an admission code.
The owner of registered planes is maintained by the administration.

The graphical constraints contained in this data model, are assigned
to object types in the following way (conforming to the style of
Elisa-D \cite{Report:91:Hofstede:LISA-D}):
\[ \footnotesize \ScriptSize \begin{array}{llll}
   C_1: & \LISA{Manufacturer}          
        & \mapsto & \LISA{TOTAL  \{  Manufacturer.builds    \} }\\
   C_2: & \LISA{Airplane}              
        & \mapsto & \LISA{UNIQUE \{ Airplane.has-as         \} }\\   
   C_3: & \LISA{Admission-code}        
        & \mapsto & \LISA{TOTAL  \{ Admission-code.given-to \} }
\end{array} \]
In this example, the expression \LISA{TOTAL \{Manufacturer.builds\}}
denotes the requirement that each \LISA{Manufacturer} plays the role
\LISA{builds}.
The expression \LISA{UNIQUE \{Airplane.has-as\}} requires uniqueness
of playing the role \LISA{has-as}.
All these constraints are owned by a single object type.
A more interesting case with respect to inheritance results by adding
the following constraint:
\begin{quote} \footnotesize
   All airplanes must have associated a manufacturer or an age.
   Unregistered airplanes must have both.
\end{quote}
The object type assignment for this constraint is:
\[ \footnotesize \ScriptSize \begin{array}{ll}
      C_4: & \LISA{Airplane} \\              
           & \mapsto \\
           & \LISA{TOTAL \{ Airplane.build-by, Airplaine.has-as \} }\Eol
      C_4: & \LISA{Registered-airplane} \\
           & \mapsto \\
           & \LISA{TOTAL \{ Airplane.build-by, Airplaine.has-as \} }\Eol
      C_4: & \LISA{Unregistered-airplane} \\
           & \mapsto \\
           & \LISA{TOTAL \{ Airplane.build-by \} AND 
                   TOTAL \{ Airplaine.has-as   \} }
\end{array} \]
The constraint $C_4$ is owned by object types \LISA{Airplane},
\LISA{Registered-airplane} and \LISA{Unregistered-airplane}.
Finally, as an illustration of a transition oriented constraint, consider
the following constraint for airplanes:
\[ \footnotesize \ScriptSize \begin{array}{ll}
      C_5: & \LISA{Airplane} \\            
           & \mapsto \\
           & \LISA{(Unregistered-airplane BEFORE Registered-airplane) 
                   EQUALS}\\
           & \Tab \LISA{Airplane}\Eol
      C_5: & \LISA{Registered-airplane} \\            
           & \mapsto \\
           & \LISA{(Unregistered-airplane BEFORE Registered-airplane)
                   EQUALS}\\
           & \Tab \LISA{Airplane}\Eol
      C_5: & \LISA{Unregistered-airplane} \\             
           & \mapsto \\ 
           & \LISA{(Unregistered-airplane BEFORE Registered-airplane)
                   EQUALS}\\
           & \Tab \LISA{Airplane}
\end{array} \]
stating that every airplane must have been unregistered before being 
a registered airplane.
The expression \LISA{$x$ BEFORE $y$} selects those instances of $x$,
which came into existance as instance of $x$,
{\em before} they became an instance of $y$.
The expression \LISA{$x$ EQUALS $y$} requires the outcome of $x$ to be equal
to the outcome of $y$.

A constraint $c$, in an application model version, will be a (usually very 
sparse) partial function $c: \OTypes \PartFunc \ConstrDef$, providing for 
every object type a {\em private} definition of the constraint.
Each modelling technique will have its own possibilities to
formulate inheritance rules,
thus governing the mapping $c$.
The domain for constraints is:
\Intro{\Constraints}{constraints}
\[ \Constraints = \OTypes \PartFunc \ConstrDef \]
Enforcing constraints on a population is discussed in the next
section.

\subsection{Methods}
The action model part of an application model version will be provided as a 
set of action specifications.
The domain for action definitions ($\TaskDef$) is determined by the
chosen modelling technique for the action model.

The, modelling technique dependent, inheritance mechanism for constraints
can be used for methods as well.
A method $m$ is regarded as a partial function $m: \OTypes \PartFunc
\TaskDef$, assigning action specifications to object types.
The set of all possible methods is the set of all these mappings:
\Intro{\Tasks}{method}
\[ \Tasks = \OTypes \PartFunc \TaskDef \]
This definition provides the formal foundation of the methods in the 
preleminary definition of the \keyword{living space} of an evolving 
information system as provided in \SRef{EIS}.

\subsection{Semantics of constraints and methods}
The semantics of both methods and constraints are defined by the relation
$\StateTrans{~}{~}$.
Therefore, we consider constraints as special methods,
as in~\cite{Report:91:Lamport:TLA}.
This leads to the following axiom:
\begin{Axiom}{AMU}
   $\ConstrDef \subseteq \TaskDef$ 
\end{Axiom}
A direct result of this axiom is: $\Constraints \subseteq \Tasks$.
Next, we focus at the semantics of methods, which are described
by $\StateTrans{~}{~}$ as transitions on
\keyword{application model histories}[application model history].
Methods are required to preserve the wellformedness properties
specified by $\IsAMH$.
\begin{Axiom}{AMU}[InvIsAMH]
   \( H \StateTrans{m}{t} H' \implies (\IsAMH(H) \implies \IsAMH(H')) \)
\end{Axiom}
The meaning of a method may depend on the history sofar of an
application model.
It may, however, not depend on any future behaviour of the application
model:
\begin{Axiom}{AMU}[PastDependence]
   \( H \StateTrans{m}{t} H'
      \implies
      H = \UpTo{H}{t} \)
\end{Axiom}
Furthermore, the effect of a method is completely known after
its completion:
\begin{Axiom}{AMU}[ImmediateResult]
   \( H \StateTrans{m}{t} H'
      \implies
      H' = \UpTo{H'}{\Inc t} \)
\end{Axiom}
The history of an application model is supposed to be monotoneous.
So it is not possible to falsify (correct) the history.
\begin{Axiom}{AMU}[ChangeNextVersionOnly]
   \( H \StateTrans{m}{t} H'
      \implies
      \UpTo{H}{t} = \UpTo{H'}{t}  \)
\end{Axiom}

Constraints are deemed as a special kind of method, behaving like a guard
on application model histories.
As a result, constraints are basically predicates.
The semantics of constraints are not influenced by the next state:
\begin{Axiom}{AMU} If $c \in \Constraints$ then
   \[ H \StateTrans{c}{t} H_1 \iff H \StateTrans{c}{t} H_2 \]
\end{Axiom}
This axiom implies that $H \StateTrans{c}{t}$ is a meaningfull expression.
\Comment{
\begin{remark}[ConstrOverPop]
   A constraint is a statement (property) about the population of an
   information structure.
   Therefore, the semantics of a constraint can be restricted to
   populations as follows:
   \begin{quote}
      If $c \in \Constraints$ then
      \[ H \StateTrans{c}{t} \iff H\Sup{\Pop} \StateTrans{c}{t} \]
      where
      \[ H\Sup{\Pop} = \Set{h \in H}{ \FuncRan(h) \subseteq \HasTypes }\]
   \end{quote} 
\end{remark} 
}

\subsection{Evolution dependency}
\SLabel{subsection}{EvolDep}

Methods (and constraints) are usually defined by some syntactic
mechanism (language).
For example,
for \SRef{\Airplanes} the specification language LISA-D is used
to express non-graphical constraints.
The graphical constraints in \SRef{\Airplanes} form another example
of the use of a (graphical!) syntactic mechanism.

Every method and constraint will refer to (uses) a
number of object types and denotable instances (i.e. directly
representable on a communication medium).
This relation is provided in the application model universe by means
of the dependency relation $\Depends$:
\Intro{\Depends}{dependency of application model elements}
\[ \Depends \subseteq (\TaskDef \union \ConstrDef) \Carth
                      (\OTypes \union \LTypes  \Carth \ConcrDoms) \]
This relation is modelling technique dependent, but is not
subject to evolution.

The interpretation of this relation is as follows:
$x \Depends y$ means that if $y$ is not alive in an application model
version, then $x$ has no meaning in that version.
A consequence is that, in case of evolution of application models, when
$y$ evolves to $y'$, then $x$ must be adapted appropriately. 

As an example, consider the second action specification from the rental 
store example:\\
~~\\
   \LISA{ACTION Init-freq} =\\
   \LISA{~~~ WHEN ADD Medium:$x$ DO}\\
   \LISA{~~~~~~ IF Lp:$x$ THEN}\\
   \LISA{~~~~~~~~~ ADD Lp:$x$ has Lending-frequency of Frequency:$0$}\\
~~\\
This action specification depends on object types \LISA{Medium, Lp} and 
\LISA{Frequency}. 
It, furthermore, depends on the domain assignment: 
\LISA{Frequency $\mapsto$ Natno}.
If one of the object types, or the domain assignment, is terminated or
changed, the action specification has to be terminated or changed accordingly. 
This will be formalized in a later section as \SRef{DanglingTypes}.

%% file: gu05.tex
\section{Application Model Versions}
\SLabel{section}{AppModVersion}

In this section, the formal definition of an application model version is
provided, containing all components from the hierarchy of models, and 
the relations among them.
First, we give a delimitation of the state space of the application model 
versions by means of an application model universe.

\input{gu0501} 
\input{gu0502} 

%% file: gu0501.tex
\subsection{Deriving Application Model Versions}

The (description of the) evolution of an \keyword{application domain}
(i.e., an \keyword{application model history}) has been introduced as a set of
\keyword{application model element evolutions}[application model element evolution].
Therefore, an \keyword{application model version} can be determined by the
actual \keyword{application model element versions}.
At this moment we will identify the domain for such versions:
\begin{definition} An application model version over application
   model universe $\AppModUniverse$ is defined as:
   \Intro{\AppMod_t}{application model version}
   \[ \AppMod_t = \tuple{\OTypes_t, \Constraints_t, \Tasks_t, \HasTypes_t,
                         \Domain_t} \]
   where $\OTypes_t      \subseteq \OTypes$,
         $\Constraints_t \subseteq \Constraints$,
         $\Tasks_t       \subseteq \Tasks$,\\
         $\HasTypes_t    \subseteq \HasTypes$ and
         $\Domain_t      \in \Domain$.
\end{definition}
From a version of an application model, we can derive the current
version $\InfStruct_t = \tuple{\LTypes_t,\NLTypes_t,\TypeRel_t,\ParentOf_t}$
of the information structure as follows:
\begin{eqnarray*}
   \LTypes_t       & =   & \OTypes_t \intersect \LTypes\\
   \NLTypes_t      & =   & \OTypes_t \intersect \NLTypes\\
   x \TypeRel_t y  & \Eq & x \TypeRel y  \land x,y \in \OTypes_t\\
   x \ParentOf_t y & \Eq & x \ParentOf y \land x,y \in \OTypes_t
\end{eqnarray*}
\Intro{\InfStruct_t}{information structure version}
As an overview of the components of an application model version, a meta 
model is provided in \SRef{\SimpleMetaMod}.
This (meta) data model is conforming to the PSM data modelling technique,
an extension of the NIAM modelling technique.
The object types $\Constraints_t$ and $\Tasks_t$ in \SRef{\SimpleMetaMod}
are power types,
the data modelling pendant of power sets in set theory.
\FiG[\SimpleMetaMod]{A meta model for information structures}

Every application model version must adhere to certain rules of well-formedness.
Some of these rules are modelling technique dependent,
and therefore outside the scope of this paper.
Nonetheless, some general rules about application model versions
can be stated.

\subsubsection{Active and living objects}
An object type $x$ is called \Keyword{alive} at a certain point of time $t$,
if it is part of the application model version at that point of time 
($x \in \OTypes_t$).
Furthermore, an object type $x$ is termed \Keyword{active} at a certain point
of time $t$,
if it is instantiated at that moment,
i.e., if there is an \Keyword{instance typing $X$ at time} $t$
such that $x \in X$.
We call $X$ an instance typing at time $t$ if
$\Ex{v,t}{v \HasTypes_t X}$.
In the remainder of this subsection,
a number a rules for instance typings will follow.

A first rule of wellformedness states that every active object type must be 
alive as well. 
This rule can be popularised as: `I am active, therefore I am alive'. 
It is formalised as:
\begin{Axiom}{AMV}[ActTypesLive](active life)
   If $X$ is an instance typing at time $t$, then:
      \[ X \subseteq \OTypes_t \]
\end{Axiom}
The next rule of wellformedness states that sharing an instance at any
point of time,
is to be interpreted as a proof of type relatedness:
\begin{Axiom}{AMV}[ActTypeRel](active relatedness)
   If $X$ is an instance typing, then:
      \[ x,y \in X \implies x \TypeRel y \]
\end{Axiom}
We call $X$ an instance typing, if $X$ is an anstance typing
at some point of time $t$.
In a later section we will prove a stronger version of this axiom.
From the very nature of the root relation it follows
that instances are included upwards, towards the roots.
As a result, every instance of an object type is also an instance of
its ancestors (if any):
\begin{Axiom}{AMV}[FoundedActivity](foundation of activity)\\
   If $X$ is an instance typing, then the relation $P$,
   defined by $P(x) = x \in X$, is a weak inheritance property.
\end{Axiom}
Applying the foundation schema (\SRef{FoundationSchema}) to
this axiom shows the presence of roots in instance typings:
\begin{lemma}[ActiveRoots](active roots)
   If $X$ is an instance typing, then:
      \[ y \in X \implies \Ex{x}{x \in X \land x \RootOf y} \]
\end{lemma}
In most traditional data modelling techniques (ER, NIAM, \ldots)
each type hierarchy has a unique root.
As a consequence,
each instance typing contains a unique root.
Some data modelling techniques (such as PSM),
however,
allow type hierarchies with multiple roots
(see \SRef{\TypeRelPropag}).
For such modelling techniques,
the following axiom guarantees a unique root for each instance typing.
\begin{Axiom}{AMV}[UniqueRoot](unique root)
   If $X$ is an instance typing and $x,y \in X$ then:
       \[ \IsRoot(x) \land \IsRoot(y) \implies x=y \]
\end{Axiom}
Some modelling techniques allow for multiple-rooted type hierarchies.
As an example, consider \SRef{\RootCS}.
The instances of object types, however, are single-rooted.
For example, we could have the following instances:
\begin{enumerate}
\item $\tuple{h_1,\set{\LISA{House},\LISA{Product},\LISA{Real estate}}}$,
\item $\tuple{b_1,\set{\LISA{Boat},\LISA{Product}}}$
\item $\tuple{b_2,\set{\LISA{Boat},\LISA{Product},\LISA{Real estate}}}$
\end{enumerate}
The instance
$\tuple{w_1,\set{\LISA{House},\LISA{Boat},\LISA{Product},\LISA{Real estate}}}$
is not a valid one,
as both \LISA{House} and \LISA{Boat} are root object types.
In this case, the value $w_1$ would inherit its identification from these
object types,
and would therefore be identified by both an adress and
boat registration number,
which is a contradiction.
The above axiom is intended to exclude such instances.

The above axiom leads to the following strengthening of \SRef{ActiveRoots}:
\begin{lemma}[RootActive](active root)
   If $X$ is an instance typing, then:
       \[ y \in X \implies \Eu{x}{x \in X \land x \RootOf y} \]
\end{lemma}
Axiom~\ref{FoundedActivity} has a structural pendant as well:
every living object type is accompanied by one of its ancestors (if any).
This is stipulated in the following axiom:
\begin{Axiom}{AMV}[LiveFoundation](foundation of live)
  The relation $P$, defined by $P(x) = x \in \OTypes_t$,
  is a weak inheritance property.
\end{Axiom}
Note that \SRef{LiveFoundation} can not be derived from
\SRef{FoundedActivity}.
The reason is that a non-root object type may be alive,
yet have no instance associated.
By applying the foundation schema on \SRef{LiveFoundation} we get:
\begin{lemma}[RootEx](living roots)
   \[ y \in \OTypes_t \implies \Ex{x}{x \in \OTypes_t \land x \RootOf y} \]
\end{lemma}
Note that in this case the root $x$ does not have to be unique.

\subsubsection{Wellformed concretisation}
In a valid application model version each label type is
\Keyword{concretised}[concretisation]
by associating a domain.
Therefore,
the domain providing function $\Domain_t$ is a (total) function
from alive label types to domains:
\begin{Axiom}{AMV}[LabDom](full concretisation)
      \( \Domain_t: \LTypes_t \Func \ConcrDoms \)
\end{Axiom}
Furthermore, the instances of label types must adhere to this domain
assignation:
\begin{Axiom}{AMV}[LabTyping](strong typing of labels)
   If $v \HasTypes_t X$ and $v \in \Union\ConcrDoms$ then:
      \[ x \in X \implies v \in \Domain_t(x) \]
\end{Axiom}

\subsubsection{Constraints and methods}
Methods, and thus constraints, are defined as mappings from object
types to method and constraint definitions respectively.
This implies that object types, owning a constraint or a method,
must be alive.
\begin{Axiom}{AMV}[LifeDef](alive definitions)
   If $w \in \Constraints_t \union \Tasks_t$ then:
      \[ \FuncDom(w) \subseteq \OTypes_t \]
\end{Axiom}
where $\FuncDom(w) = \Set{x}{\tuple{x,y} \in w}$ is the domain of
function $w$.
For example, constraint $C_1$ from the airplane example can only
be alive if the object type \LISA{Manufacturer} is alive.
As a next rule, object types that own the same constraint or method,
must be type related.
\begin{Axiom}{AMV}[TypeRelDef](type related definitions)\\
   If $w \in \Constraints_t \union \Tasks_t$ then:
      \[ x,y \in \FuncDom(w) \implies x \TypeRel y \]
\end{Axiom}
For example, object types \LISA{Registered-airplane} and
\LISA{Unregistered-airplane} both own constraint $C_4$.
As a result, $C_4$ constrains populations of both object types.
This only makes sense if the object types \LISA{Registered-airplane} and
\LISA{Unregistered-airplane} can have values in common, i.e., if they
are type related.

Finally, due to inheritance, if a constraint is defined for an ancestor
object type, it is defined for all its offspring as well.
For example, if a constraint puts a limitation on airplanes,
then this constraint is also effective for special kinds of airplanes
such as registered airplanes.
\begin{Axiom}{AMV}[DefInheritance](inheritance of definitions)\\
   If $w \in \Constraints_t \union \Tasks_t$ then
   the relation $P$, defined by $P(x) = x \in \FuncDom(w)$,
   is a strong inheritance property.
\end{Axiom}
Note that the inheritance direction for populations, is reverse to the
inheritance direction for methods (and constraints).
         
The motivation for the next axiom lies in the following observation
(see \SRef{EvolDep}).
The definition of a constraint or a method refers to a set of object types,
and domain concretisations.
Thus, if a method or constraint definition is alive, then all these
referred items should be alive at that same moment.
\begin{Axiom}{AMV}[DanglingTypes](dangling references)\\
   If $w \in \Constraints_t \union \Tasks_t$ then:
      \[ w(x) \Depends y \implies
         y \in \OTypes_t \union (\LTypes_t \Carth \ConcrDoms_t)\]
\end{Axiom}
Since every instance from a non-root object type is inherited downwards
in the identification hierarchy towards the root object types, constraints
on child-object types should be at least as restrictive:
\begin{Axiom}{AMV}(strengthening of constraints)[AMVStrength]
   If $c \in \Constraints_t$, then:
   \[ x \ParentOf y \land \DefinedAt{c}{x,y} \implies
      c(y) \ConstrImplies{\:} c(x) \]
\end{Axiom}
   where $d_1 \ConstrImplies{\:} d_2$ is defined as:
   \[ d_1 \ConstrImplies{\:} d_2 ~\Eq~
      \Al{t,H}{ H \StateTrans{d_1}{t} \implies H \StateTrans{d_2}{t} }
   \]
   The intuitive meaning of $d_1 \ConstrImplies{\:} d_2$
   is: $d_1$ is at least as restrictive as $d_2$ (see also
   \cite{PhdThesis:92:Hofstede:DataMod}).
As an illustration of this rule, consider constraint $C_4$ of
the airplane example.
For unregistered airplanes is a strengthening of the rule for airplanes.
It would not make sense to be more liberal for unregistered
airplanes than for airplanes in general,
as each unregistered airplane is also an airplane!

%% file: gu0502.tex
\subsection{Populations of information structures}
\SLabel{subsection}{Pops}

A special part of an application model version is its population.
This population can be derived from the relation $\HasTypes_t$:
\begin{definition}
   The population of an information structure at any point of time,
   is a mapping
   $\Pop: \Time \Func (\OTypes \Func \Powerset(\UniDom))$,
   defined by:
   \[ \Pop_t(x) = \Set{v}{\Ex{Y}{v \HasTypes_t Y \land x \in Y}} \]
   \Intro{\Pop_t}{population version}
\end{definition}
It will be convenient to have an overview of all instances that ever lived.
We will refer to this population as the extra-temporal population.
\begin{definition}
   The extra-temporal population of an application model is a mapping
   $\Pop_\infty: \OTypes \Func \Powerset(\UniDom)$,
   defined by
   \[ \Pop_\infty(x) = \bigcup\Sub{t \in \Time}\Pop_t(x) \]
   \Intro{\Pop_\infty}{extra-temporal-population}
\end{definition}
Axiom~\ref{FoundedActivity} relates instances to the object type hierarchy.
This leads to the following property for populations:
\begin{lemma}[PopDist](population distribution)
   Every instance of an object type, is also instance of one of
   its roots:
   \[ \Pop_t(x) \subseteq \bigcup\Sub{y \RootOf x} \Pop_t(y) \]
\end{lemma}
\begin{proof}
   Let $i \in \Pop_t(x)$ then:

   $i \in \Pop_t(x)$\\
   $\EqTo$ \Explain{definition of $\Pop_t$}

   $\Ex{X}{i \HasTypes_t X \land x \in X}$\\ 
   $\Therefore$ \Explain{\PSRef{ActiveRoots}}

   $\Ex{y,X}{i \HasTypes_t X \land x,y \in X \land y \RootOf x}$\\
   $\EqTo$ \Explain{definition of $\Pop_t$}

   $\Ex{y}{y \RootOf x \land i \in \Pop_t(y)}$\\
   $\EqTo$ \Explain{elaborate}

   $i \in \bigcup\Sub{y \RootOf x} \Pop_t(y)$.
\end{proof}
The result of the previous lemma can be generalised to extra-temporal 
populations:
\begin{corollary}
   \[ \Pop_\infty(x) \subseteq \bigcup\Sub{y \RootOf x} \Pop_\infty(y) \]
\end{corollary}
\begin{proof}
   Let $v \in \Pop_\infty(x)$ then:

   $v \in \Pop_\infty(x)$\\
   $\EqTo$ \Explain{definition of $\Pop_\infty$}

   $\Ex{t}{v \in \Pop_t(x)}$\\
   $\Therefore$ \Explain{\PSRef{PopDist}}

   $\Ex{t}{v \in \bigcup\Sub{y \RootOf x} \Pop_t(y)}$\\
   $\EqTo$ \Explain{definition of $\Pop_\infty$} 
   
   $v \in \bigcup\Sub{y \RootOf x} \Pop_\infty(y)$
\end{proof}
Next we focus at strong typing,
which is considered to be a property to hold on each moment:
if $x \not\TypeRel y$, then their populations may never share instances.
The following axiom is sufficient to guarantee this property,
as we will show in \SRef{StrongTyping}.
\begin{Axiom}{AMV}[ExclRootPop](exclusive root population)
   If $\IsRoot(x)$ and $\IsRoot(y)$ then:
      \[ x \not\TypeRel y \implies 
         \Pop_\infty(x) \intersect \Pop_\infty(y) = \emptyset \]

   If roots are not type related,
   then their extra-temporal populations are disjoint.
\end{Axiom}
By means of the following theorem the nature of type relatedness, captured 
for roots in the above axiom, is generalised to object types in general:
\begin{theorem}[ExclPop](exclusive population)
   If $x \not\TypeRel y$ then 
   \[ \bigcup\Sub{z \RootOf x}\Pop_\infty(z) ~\intersect~ 
      \bigcup\Sub{z \RootOf y}\Pop_\infty(z) ~=~ \emptyset \]
   The populations of object types which are not type related, have
   no values in common.
\end{theorem}
\begin{proof}
   $x \not\TypeRel y$\\
   $\EqTo$ \Explain{\PSRef{TypeRelPropagation}}

   $\neg\Ex{z_1,z_2}{%
       z_1 \RootOf x \land z_2 \RootOf y \land
       z_1 \TypeRel z_2 
   }$\\
   $\Therefore$ \Explain{elaborate}

   $\Al{z_1,z_2}{%
       z_1 \RootOf x \land z_2 \RootOf y \implies
       z_1 \not\TypeRel z_2 
   }$\\ 
   $\Therefore$ \Explain{\SRef{ExclRootPop}}

   $\Al{z_1,z_2}{
       \begin{array}{l}
          z_1 \RootOf x \land z_2 \RootOf y \implies \\
          \Tab \Pop_\infty(z_1) \intersect \Pop_\infty(z_2) = \emptyset 
       \end{array}
   }$\\
   $\Therefore$ \Explain{elaborate}

   $\bigcup\Sub{z \RootOf x}\Pop_\infty(z) \intersect 
    \bigcup\Sub{z \RootOf y}\Pop_\infty(z) = \emptyset $
\end{proof}
From \SRef{PopDist} and \SRef{ExclPop} the main typing
theorem is derived:
\begin{theorem}[StrongTyping](strong typing theorem)
   \[ x \not\TypeRel y \implies 
      \Pop_\infty(x) \intersect \Pop_\infty(y) = \emptyset \]
\end{theorem}

We are now in a position to define what constitutes a wellformed application
model version.
Let $\AppMod_t=\tuple{\OTypes_t, \Constraints_t, \Tasks_t,
                   \HasTypes_t, \Domain_t}$:
\begin{eqnarray*}
   \IsAM(\AppMod_t) & \Eq & \IsSch(\OTypes_t) \land 
                            \IsPop(\OTypes_t,\HasTypes_t) \land\\
                    &     & \AppMod_t 
                            \mbox{adheres to the \ClassRef{AMV} axioms}
\end{eqnarray*}
In the next section, this predicate will be used to define what constitues a
proper application model history ($\IsAMH$).

%% file: hu06.tex
\section{Evolution of Application Models}
\SLabel{section}{AMEvolv}

As stated before, the evolution of an application model is described
by the evolution of its elements.
The set $\AME$ was introduced as the set of all evolvable elements of an
application model.
Its formal definition in terms of components of $\AppModUniverse$ is:
\begin{definition} Application model elements:
   \Intro{\AME}{application model elements}
   \[
      \AME = \OTypes \union \Constraints \union \Tasks \union
             \HasTypes \union \Domain
   \]
\end{definition}
An \keyword{application model element evolution} was defined as a partial function,
assigning actual version of application model elements to points of
time.
Note that the type relatedness and root relationships are defined for the
evolution state space as a whole, and are therefore not subject to any
evolution.

In this section we will present a set of wellformedness rules
for application model histories.
These rules represent our {\em way of thinking} with regards to a
wellformed evolution, which is based on strong typing and a strict
notion of identification of instances.
Alternative {\em ways of thinking}, and corresponding wellformedness
rules may be chosen.
For the remainder of this section,
let $H$ be some (fixed) application model history.

\input{gu0601} 
\input{gu0602} 
\input{gu0603} 
\input{gu0604} 
\input{gu0605} 
\input{hu0606} 

%% file: gu0601.tex
\subsection{Separation of element evolution}
\SLabel{subsection}{SepElems}

The first rule of wellformedness states that
the evolution of application model elements is bound to element classes.
For example,
an object type may not evolve into a method,
and a constraint may not evolve into an instance.
The motivation behind this rule is strong typing at a theory level.
Usually, strong typing leads to better structured models, while type
checking provides a means for error detection.
This is formalised in the following axiom:
\begin{Axiom}{EW}(evolution separation)\\
   If $X \in \set{\OTypes, \Constraints,  \Tasks, \HasTypes, \Domain}$,
   and $h \in H$ then:
   \[ h(t) \in X \implies \FuncRan(h) \subseteq X \]
   where $\FuncRan(h) = \Set{y}{\tuple{x,y} \in h}$.
\end{Axiom}
From this axiom it follows that an application model history can be
partitioned into the history of its object types, its constraints,
its methods, 
its populations, and its concretisations (of label types):
\begin{definition}
   object type histories:\\
   \Tab \Tab 
   $\ObjEvol = \Set{h \in H}{\exists_t[h(t) \in \OTypes]}$\Eol
   \Intro{\ObjEvol}{object type histories}
   constraint histories:\\ 
   \Tab \Tab  
   $\ConstrEvol =  \Set{c \in H}{\exists_t[c(t) \in \Constraints]}$\Eol
   \Intro{\ConstrEvol}{constraint histories}
   method histories:\\
   \Tab \Tab 
   $\TaskEvol = \Set{m \in H}{\exists_t[m(t) \in \Tasks]}$\Eol
   \Intro{\TaskEvol}{method histories}
   population histories:\\
   \Tab \Tab 
   $\PopEvol = \Set{g \in H}{\exists_t[g(t) \in \HasTypes]}$\Eol
   \Intro{\PopEvol}{population histories}
   concretisation histories:\\
   \Tab \Tab 
   $\DomEvol = \Set{d \in H}{\exists_t[d(t) \in \Domain]}$
   \Intro{\DomEvol}{concretisation histories}
\end{definition}
In \SRef{AppMod}, an application model version was introduced
($\AppMod_t$) as the following tuple:
\[ \AppMod_t =
   \tuple{\OTypes_t, \Constraints_t,  \Tasks_t, \HasTypes_t, \Domain_t} \]

%% file: gu0602.tex
\subsection{Deriving application model versions}
An application model version
$\AppMod_t(H)$
$=$
$ \tuple{\OTypes_t, \Constraints_t,  \Tasks_t, \HasTypes_t, \Domain_t}$
at a given point of time $t$, is easily
derived from an application model history $H$.
This is done by defining the five main components, which determine an
application version:
\begin{definition}[AMVersion]
   object types:\\
   \Tab \Tab 
   $\OTypes_t = \Set{h(t)}{h \in \ObjEvol \land \DefinedAt{h}{t}}$\Eol
   constraints:\\
   \Tab \Tab 
   $\Constraints_t = \Set{c(t)}{c \in \ConstrEvol \land \DefinedAt{c}{t}}$\Eol
   methods:\\
   \Tab \Tab 
   $\Tasks_t = \Set{m(t)}{m \in \TaskEvol \land \DefinedAt{m}{t}}$\Eol
   population:\\
   \Tab \Tab 
   $\HasTypes_t = \Set{g(t)}{g \in \PopEvol \land \DefinedAt{g}{t}}$\Eol
   concretisations:\\
   \Tab \Tab 
   $\Domain_t = \Set{d(t)}{d \in \DomEvol \land \DefinedAt{d}{t}}$
\end{definition}
In this definition $\DefinedAt{f}{t}$ is used as an abbreviation
for $\Ex{s}{\tuple{t,s} \in f}$,
stating that (partial) function $f$ is defined at time $t$.

\FiG[\EvolutieMetaMod]{A meta model for the evolution system}
As an outline of the hitherto defined concepts, a (meta) data model, 
relating all defined concepts, is provided in \SRef{\EvolutieMetaMod}.
The data model depicted there is conform the PSM modelling technique,
and uses the notion of schema objectification (object type $\AMH$),
and power typing (object type $\Powerset(\OTypes)$).
The population of an objectified schema at hand is to be looked upon as one 
single abstract object instance of the object type corresponding to the
objectified schema.
Power-typing is, as stated before, the data modelling pendant of power-sets
from set theory.

%% file: gu0603.tex
\subsection{Enforcing constraints}
   
As a next rule of well-formedness on the evolution of an application model
history $H$,
the following axiom states that all constraints must hold:
\begin{Axiom}{EW}[ConstraintsHold](constraints hold)
   For all $c \in \ConstrEvol$:
   \[ \DefinedAt{c}{t} \implies H[T] \StateTrans{c(t)}{t} \]
   where $T$ is the largest time interval such that:
   \[ \forall_{t' \in T}[t' \leq t \land c(t')=c(t)] \]
   and furthermore:
   \[ H[T] = \Set{h[T]}{h \in H} \]
\end{Axiom}
Note that the constraint $c$ is only enforced for the population
valid during the validity of the constraint itself.


\begin{remark}
   In programming, it is considered bad practice to write
   \Keyword{self modifying code}, i.e. a program that modifies itself.
   Analogously, one could formulate a rule stating that actions in an
   application model should only have effect on the population.
   Any structural changes (actions, information structure, \ldots) should thus 
   be performed by an action specified by the user explicitly (an update 
   request).
   This can be formulated as:
   \begin{quote}
      If $m \in \Tasks_t$ then
      \[ H_t           \StateTrans{m}{t} H_{\Inc t} \iff
         H\Sup{\Pop}_t \StateTrans{m}{t} H\Sup{\Pop}_{\Inc t} \]
      The $H\Sup{\Pop}$ is the restriction of $H$ to the evolution of the
      elements of $\HasTypes$.
      \Comment{, as used in \SRef{ConstrOverPop} as well.}
   \end{quote}
\end{remark}

%% file: gu0604.tex
\subsection{An example of element evolution}
As an example of evolution, the following table respresents three
object type evolutions ($h_1,h_2,h_3 \in \ObjEvol$) from the rental store
(see \SRef{Approach}):
\[ \footnotesize \SmallSize \begin{array}{c|ccc}
   \ObjEvol & h_1 & h_2 & h_3 \\
   \hline
   t_1 & o_1 & o_2 & - \\
   t_2 & o_1 & o_2 & - \\
   t_3 & o_1 & o_2 & - \\
   t_4 & o_1 & o_3 & o_4 \\
   t_5 & o_1 & o_3 & o_4
\end{array} \]
where $\Vector{t}{5} \in \Time$ , and $\Vector{o}{4} \in \OTypes$ are
object types, such that:
\[ \footnotesize \SmallSize
  \begin{array}{lcl}
    o_1 & = & \mbox{`Entity type: \LISA{Record}'}  \\
    o_2 & = & \mbox{`Fact type:   \LISA{Recording of Song on Record}'} \\
    o_3 & = & \mbox{`Fact type:   \LISA{Recording of Song on Medium}'} \\
    o_4 & = & \mbox{Entity type:  \LISA{Medium}'}
  \end{array}
\]
Note that the evolution step (from \SRef{\LPLibrary} to \SRef{\LPCDLibrary})
takes place at point of time $t_4$.
Two example instance evolutions ($g_1,g_2 \in \PopEvol$), obeying the
above schema evolution, are:
\[ \footnotesize \begin{array}{c|c|c}
   \PopEvol & g_1 & g_2 \\
   \hline
   t_1  & \tuple{i_1,\set{o_1}}       & \tuple{i_2,\set{o_2}} \Forc{3} \\
   t_2  & \tuple{i_1,\set{o_1}}       & \tuple{i_2,\set{o_2}} \Forc{3} \\
   t_3  & \tuple{i_1,\set{o_1}}       & \tuple{i_3,\set{o_2}} \Forc{3} \\
   t_4  & \tuple{i_1,\set{o_1,o_4}}   & \tuple{i_4,\set{o_3}} \Forc{3} \\
   t_5  & \tuple{i_1,\set{o_1,o_4}}   & \tuple{i_4,\set{o_3}} \Forc{3}
\end{array} \]
where $\Vector{i}{4} \in \UniDom$ are instances such that:
\[ \footnotesize
  \begin{array}{lcl}
    i_1 & = & \mbox{`Brothers in Arms'} \Eol
    i_2 & = & \tuple{\mbox{`Money for nothing'},\mbox{`Brothers in Arms'}} \Eol
    i_3 & = & \tuple{\mbox{`Brothers in Arms'}, \mbox{`Brothers in Arms'}} \Eol
    i_4 & = & \tuple{\mbox{`Brothers in Arms'}, \mbox{`Brothers in Arms'}}
  \end{array}
\]
The interpretation of this table leads to:
\begin{description} \footnotesize \SmallSize
\item[$g_1(t_4)=\tuple{i_1,\set{o_1,o_4}}$] means:
   `Brothers in Arms' is both a \LISA{Record} and a \LISA{Lp} at $t_4$,
\item[$g_2(t_3)=\tuple{i_3,\set{o_2}}$] means:
   \LISA{Song} `Brothers in Arms'
   \LISA{is} \LISA{Recorded} \LISA{on} \LISA{Record} `Brothers in Arms',
\item[$g_2(t_4)=\tuple{i_4,\set{o_3}}$] means:
   \LISA{Song} `Brothers in Arms'
   \LISA{is} \LISA{Recorded} \LISA{on} \LISA{Medium} `Brothers in Arms'.
\end{description}

%% file: gu0605.tex
\subsection{Evolution of the identification hierarchy}
\SLabel{subsection}{EvolvIdentHier}

Thus far we discussed the wellformedness of the evolution of
application model elements.
However, as a result of object type evolution,
the identification hierarchy will evolve as well.
This evolution is not completely free, some conservatism with respect
to such evolution is appropriate.
The motivation of this approach is our tendancy to strong typing
and strict object identification.
In the remainder of this section,
we provide some rules which exclude undesirable evolutions.
It should be stressed that attacking the wellformedness problem
from another vantage-point may result in other rules.
 
Firstly,
the order in the identification hierarchy should not change in
one step, since this could lead to conflicting identification schemas
in the course of time:
\begin{Axiom}{EW}[MonotonousParents](monotonous ancestors)
   If $h_1,h_2 \in \ObjEvol$,
      $\DefinedAt{h_1}{t}$, $\DefinedAt{h_2}{t}$,
      $\DefinedAt{h_1}{\Inc t}$ and $\DefinedAt{h_2}{\Inc t}$ then
   \[ h_1(t) \ParentOf h_2(t) \land  h_1(\Inc t) \TypeRel h_2(\Inc t)
      \implies 
      h_1(\Inc t) \ParentOf h_2(\Inc t) \]
\end{Axiom}
In the CD store running example, when CD's are a special kind of Medium,
the reversal of this relation in one step is excluded by this rule,
as this would lead to identification problems for LP's.
In the airplane example,
registered airplanes are identified as airplanes in general.
Suppose registered airplanes need an identification of their own.
Then this is only possible after breaking the type relatedness
between both object types, i.e.,
breaking up the identification hierarchy.

This is not only true at the type level, but also at the evolutionary level.
A direct consequence of this axiom is that all ancestors of an object type
have to be terminated when this object type is promoted to be
a root object type:
\begin{lemma}
   If $h_1,h_2 \in \ObjEvol$,
      $\DefinedAt{h_1}{t}$, $\DefinedAt{h_2}{t}$ 
      and $\DefinedAt{h_2}{\Inc t}$ then
   \( h_1(t) \ParentOf h_2(t) \land  h_1(\Inc t) \TypeRel h_2(\Inc t)
      \land \IsRoot(h_2(\Inc t))
      \implies
      \neg \DefinedAt{h_1}{\Inc t} \)
\end{lemma}
\begin{proof}
   Let $h_1(t) \ParentOf h_2(t) \land  h_1(\Inc t) \TypeRel h_2(\Inc t) \land \IsRoot(h_2(\Inc t))$
   and $\DefinedAt{h_1}{\Inc t}$, then:

   $h_1(t) \ParentOf h_2(t) \land  h_1(\Inc t) \TypeRel h_2(\Inc t) \land \IsRoot(h_2(\Inc t)) \land 
    \DefinedAt{h_1}{\Inc t}$\\
   $\Therefore$ \Explain{\SRef{MonotonousParents}}

   $h_1(\Inc t) \ParentOf h_2(\Inc t) \land  h_1(\Inc t) \TypeRel h_2(\Inc t) \land \IsRoot(h_2(\Inc t))$\\
   $\Therefore$ \Explain{definition of $\IsRoot$}

   $\Contradiction$
\end{proof}
The following rule for identification hierarchy evolution states
that the type-instance relation (derived from the relation $\HasTypes$)
is to be maintained in the course of evolution.
Like the previous rule, the motivation of this rule is to prevent
conflicting identification schemas in the course of time.
This leads to the axiom of guided evolution:
\begin{Axiom}{EW}(guided evolution)
   If $g \in \PopEvol$,
      $\DefinedAt{g}{t}$ and $\DefinedAt{g}{\Inc t}$ then
   \[ \Ex{h \in \ObjEvol}{
          h(t) \TypeRel \ActiveAt{g}{t} \implies
          h(\Inc t) \TypeRel \ActiveAt{g}{\Inc t} } \]
\end{Axiom}
where $x \TypeRel Y$ is defined as $\Ex{y \in Y}{x \TypeRel y}$.
The types that are associated with an instance evolution $g$,
at point of time $t$,
are introduced by:
\[ \ActiveAt{g}{t} \Eq \Union\Sub{X: g \HasTypes_t X} X \]
As an example, consider the evolution of registered airplanes to an
object type with its own identification, within a separate identification
hierarchy.
Then it would not make any sense if the instances of this object type
would not follow this evolution step,
the only exception being instances that violate newly introduced constraints.
This latter aspect will be elaborated further in the next subsection.
Finally, we can live up to our promise of defining $\IsAMH$ formally:
\begin{definition}
\begin{eqnarray*}
   \IsAMH(H)
   & \Eq   & \Al{t \in \Time}{\IsAM(\AppMod_t)}\\
   & \land & \mbox{$H$ adheres to the $\ClassRef{EW}$ axioms}
\end{eqnarray*}
\end{definition}

%% file: hu0606.tex
\subsection{Propagating modifications}

When an element of the application model evolves (is modified),
other elements may have to be modified
accordingly as these modifications may invalidate others or may 
result in conflicts.
For instance, when the subtyping of object type Medium is terminated
in the LP and CD store running example, all its subtypes must be
terminated as well.
Even more, any relationship type in which such a subtype is involved
must be modified or terminated within the same transaction.

Other dependencies can be found, for example in the context of constraints.
Whenever a new constraint is added, existing instances may be in conflict
with this new rule, and must be adopted to meet the new 
requirements within the same transaction.

These dependencies are enforced on application model histories by
the relations $\IsSch$, $\IsPop$, and $\Depends$,
which require at each point in time the population (at that moment) to be
in accordance with the information structure version (at that moment).
Besides, the information structure version should satisfy the
wellformedness rules of the underlying data modelling technique.
A detailed discussion of propagation of dependencies can only be
given in the context of an application to a concrete modelling
technique.
When doing so, the issues concerning propagation of changes as discussed
in e.g.~\cite{Article:86:Skarra:OOEvolution}, 
\cite{Article:87:Banerjee:OOSchemaEvol} come into play.
For more details of the propagation of dependencies in the context of
some applications of the general theory to existing modelling
techniques,
refer to~\cite{Report:93:Proper:EvolvPSM} or
\cite{PhdThesis:94:Proper:EvolvConcModels}.

%% file: hu07.tex
\section{Conclusions and Further research}

In this paper we presented a first attempt to a general theory for the
evolution of application models, supporting evolving information systems.
In order to validate the theory, it must be applied to some modelling
techniques.

In the mean time the theory has been applied to PSM,
resulting in EVORM (\cite{Report:93:Proper:EvolvPSM},
\cite{PhdThesis:94:Proper:EvolvConcModels}), and the
conceptual transaction modelling technique Hydra
(\cite{Report:91:Hofstede:TaskAlg},\cite{PhdThesis:92:Hofstede:DataMod}),
leading to Hydrae (\cite{PhdThesis:94:Proper:EvolvConcModels}).

Furthermore, based on the notion of evolution as laid down in the axioms of
the general theory, a query and manipulation language has been defined
supporting the evolution of information systems, and disclosure of information
in an evolving context (\cite{Report:93:Proper:DisclSch},
\cite{Report:93:Hofstede:DisclSupport}).
Query formulation in the context of an evolving information system
poses extra requirements for the query language and mechanisms used
to formulate the queries, since the underlying conceptual schema
evolves in the course of time, and data stored in the old
schemas must be retrievable as well.

Remaining issues for further research are the implementation of an
actual evolving information system, the development of an adequate
modelling procedure to cope with evolution of the universe of discourse
and reflect these correctly in the information system,
Finally, the consequences of evolution for the internal representation of
information structures should be studied in more detail.

%% file: hu08.tex
\section*{Acknowledgements}

The investigations were partly supported by the 
Foundation for Computer Science in the Netherlands (SION) 
with financial support from the Dutch Organization 
for Scientific Research (NWO).

We would like to thank the anonymous referees for their many valuable
comments on earlier versions of this paper.